%% file: iclr2027_conference.tex
\documentclass{article} % For LaTeX2e
\usepackage[preprint]{diga}

\input{math_commands.tex}

\usepackage{graphicx}
\usepackage{placeins}
\usepackage{hyperref}
\usepackage{url}
\usepackage{booktabs}
\usepackage{wrapfig}
\usepackage{enumitem}
\usepackage{float}
\usepackage{lineno}

\hypersetup{colorlinks=true, citecolor=SMUGold, linkcolor=SMUBlue, urlcolor=SMUBlue}

\title{AlphaIdeaBench: Benchmarking Automated \\ Research Idea Generation under Feasibility \\ Constraints with Peer Review Alignment}

\author{Shunchao Zhou\textsuperscript{1},
    Chang Yang\textsuperscript{2},
    Jingrui Wang\textsuperscript{1,3},
    Shuxin Zhuang\textsuperscript{4},\\
    \textbf{Rundong Wang}\textsuperscript{5},
    \textbf{Runsheng Yu}\textsuperscript{6},
    \textbf{Xinrun Wang}\textsuperscript{1}\thanks{Xinrun Wang is the corresponding author.}
    \\
    \textsuperscript{1}Singapore Management University,
    \textsuperscript{2}The Hong Kong Polytechnic University\\
    \textsuperscript{3}Hangzhou Dianzi University,
    \textsuperscript{4}City University of Hong Kong\\
    \textsuperscript{5}NetEase, Inc.,
    \textsuperscript{6}The Hong Kong University of Science and Technology
    \\
    sc.zhou.2025@phdcs.smu.edu.sg, xrwang@smu.edu.sg
}

\begin{document}

\ifsubmission
\linenumbers
\fi

\maketitle

\begin{abstract}

Research ideas are always proposed under resource constraints: a plan must fit the available compute, API credit, and time. Yet evaluations of research-ideation agents mostly score unconstrained novelty or overlap with a published paper, and their LLM judges are rarely checked against peer review. We introduce AlphaIdeaBench, a benchmark of 100 ideation tasks (50 workshop calls for papers and 50 open problems stated in surveys), each paired with an explicit resource budget and a shared proposal schema. An LLM judge scores each proposal on 12 dimensions covering validity, novelty, and feasibility. To validate the judge, we compare its scores on four novelty axes (problem, method, scenario, and insight) with scores derived from reviewer comments on 600 papers from ICLR 2026, NeurIPS 2025, and ICML 2025. For problem and method novelty, Gemini 3.7 Flash is within one point of the reviewer-derived score on 94–97\% of papers; agreement on scenario novelty is relatively lower. We evaluate eight published harnesses on a shared backbone and, within one harness, eight backbone LLMs. CoI-Agent obtains the highest overall score, narrowly ahead of SciMON, which leads on method and insight novelty. All harnesses score low on scenario novelty and rarely change the problem setting. The backbone largely determines which open problem a proposal targets, whereas the harness determines whether a proposal addresses one problem or combines several. Weaker backbones such as Llama 4 Scout fail to carry out that harness's strategy and score low on budget feasibility. Finally, when only the budget is varied on three topics, a larger budget changes the experimental setting on compute-intensive topics but only adds tests on an inexpensive one. We release the tasks, rubrics, and scores  to foster standardized, reproducible benchmarking  of future research-ideation agents under constraints.
\end{abstract}

\section{Introduction}

Automating research is no longer only a writing aid. Agents now read a literature, name an open question, and draft the experiment that later code and papers will carry out \citep{lu2024ai,yamada2025ai,baek2025researchagent,wang2024scimon,li2025coi,su2025virtualscientists}. That first write is the expensive step. A plan that is fluent but empty still consumes GPU time, API credit, calendar, and reviewer attention, and systems that continue into execution or manuscript drafting cannot repair a weak question \citep{schmidgall2025agentlaboratory,tang2025airesearcher}.If automated research is going to be used, the idea itself has to be something a group could realistically defend and actually run.

Producing that idea is harder than producing text that looks like a paper. A usable proposal has to stay inside the given literature, be novel in the question, the method, or the setting in which the method is run, and fit a budget the authors can actually spend. These properties do not move together. Overlap with one published paper punishes a legitimate departure from that paper, and a single overall score hides which of them failed. Current comparisons do not separate them. Each harness, the scaffold of retrieval, critique, and rewrite steps wrapped around a language model, is usually shown with its own prompt and its own judge, and with no shared statement of how much compute or calendar the author may spend \citep{si2024can,chen2025mlrbench}. Studies that ask humans whether a model idea matches an expert idea do not leave a reusable checklist. Language-model judges scale, but without a check against peer review they are easy to overtrust \citep{zheng2023judging,liu2023geval}. Workshop suites such as {MLR-Bench} package a real call for papers, then move on to later experimental and writing stages \citep{chen2025mlrbench}. What is still missing is a benchmark of the written idea: one context, one resource envelope, one schema, and a novelty judge checked against reviewer text from premier machine learning conferences.

AlphaIdeaBench is that testbed. Every task file gives the agent the same scholarly context, the same resource envelope, and the same JSON schema (title, question, hypothesis, method, experiments, baselines, success criteria, related work, limitations, and a resource plan). We score the completed proposal on 12 dimensions covering validity, novelty, and feasibility, and we release the checklists. Novelty is split into problem, method, scenario, and insight. Before those axes are used on agent writes, we map them from reviewer comments on 600 papers at ICLR 2026, NeurIPS 2025, and ICML 2025 and compare the judge with that mapping. The tasks follow the two routes a human researcher actually uses: 50 workshop calls and 50 open problems taken from surveys. In the main study the envelope is an academic lab budget: four A100 GPUs, 200 GPU hours, \$200 of API credit, eight weeks, and 40 annotation hours. On the 600 papers, problem and method novelty from the Gemini 3.7 Flash judge sit within one point of the reviewer-mapped scores on 94\% to 97\% of items. On the 100 tasks, eight harnesses that share this strong backbone fill validity and feasibility, so their rank is mostly a novelty rank. That ceiling depends on the model: under the same harness, a weak backbone loses validity and feasibility as well. {CoI-Agent} is best overall (4.31) because it pairs novelty with near-ceiling testability. Scenario novelty stays low for every harness, at most 1.87. Holding the topics fixed, the pretraining prior of the model decides which listed challenge looks worth writing, while the harness decides whether to dig one bottleneck or to fuse several. Stretching the same envelope from a student budget to a lab budget, extra compute changes the experiment on costly topics, moving from simulation toward the real system, and is spent as more tests of the same plan when the topic is already cheap to run.
Our contributions are:
\begin{itemize}[leftmargin=*, topsep=0pt, partopsep=0pt, parsep=0pt, itemsep=0pt]
    \item A public 12-dimension scoring rubric, and released fine-grained novelty scores (problem, method, scenario, and insight) mapped from reviewer comments on 600 ICLR, NeurIPS, and ICML papers.

    \item A 100-task suite that abstracts two human routes to an idea, reading a survey and answering a workshop CFP, with 50 tasks in each route and an explicit resource cap.

    \item A factorial comparison that separates the harness from the ideation model: eight published harnesses on one backbone, and eight backbones under one harness.
    % , i.e.\ the agent's style of thinking and working.

    \item An envelope study that varies only the budget, on topics whose experiments differ in cost, and asks how the proposed idea changes in ambition and experimental scope.
\end{itemize}

\section{Related work}
\label{sec:related_work}

\textbf{Autonomous Scientific Agents.}
End-to-end systems try to run more than ideation. {AI-Scientist-V1} generates hypotheses, runs experiments from code templates, and drafts a paper \citep{lu2024ai}. {AI-Scientist-V2} drops those templates and uses agentic tree search over experiments \citep{yamada2025ai}. {Agent Laboratory} takes a human-provided idea and moves it through literature review, experimentation, and report writing with PhD, postdoc, and engineer roles \citep{schmidgall2025agentlaboratory}. {AI-Researcher} orchestrates literature review, hypothesis generation, implementation, and manuscript writing; Scientist-Bench is the evaluation suite introduced in that paper, not the agent itself \citep{tang2025airesearcher}. Other lines target the ideation step. {ResearchAgent} retrieves related papers and knowledge entities, then iteratively refines the problem, method, and experiment design with LLM reviewers \citep{baek2025researchagent}. {CoI-Agent} organizes retrieved papers into a progressive idea chain rather than a bag of citations, then writes an idea and an experiment plan \citep{li2025coi}. {SciMON} retrieves inspirations from prior work and rewrites the suggestion until a novelty comparison with retrieved papers is cleared \citep{wang2024scimon}. {VirSci} (Virtual Scientists) forms a team from author profiles and runs inter- and intra-team discussions to produce an idea \citep{su2025virtualscientists}. These systems are usually demonstrated with their own prompts and judges, and several originally continue into code execution or manuscript writing. AlphaIdeaBench holds the topic, resource envelope, output schema, and judge fixed, and scores only the final written research proposal.

\textbf{Benchmarks for Scientific Ideation.}
Scoring a research idea is not one number: novelty, soundness, and whether the plan can be run do not move together. Early automatic scores use overlap with a gold paper and punish legitimate departures. Human studies ask whether models can match expert ideas \citep{si2024can}, but leave no reusable checklist. {VBench} makes the same point for video, splitting one quality score into separate dimensions \citep{huang2024vbench}; we take that split as the right grain for ideation. {NovBench} checks LLM novelty assessments against reviewer text on published papers \citep{wu2026novbench}. We likewise check four novelty axes against reviewer comments, rather than treating one overall score as a novelty label. {MLR-Bench} packages workshop CFPs and then scores later experiments and writing \citep{chen2025mlrbench}. We stop at the  proposal and score both its novelty and feasibility under one shared, explicit academic resource envelope.

\section{The AlphaIdeaBench benchmark}
\label{sec:framework}

\begin{figure}[t]
% \begin{center}
\vspace{-15pt}
\includegraphics[width=\linewidth]{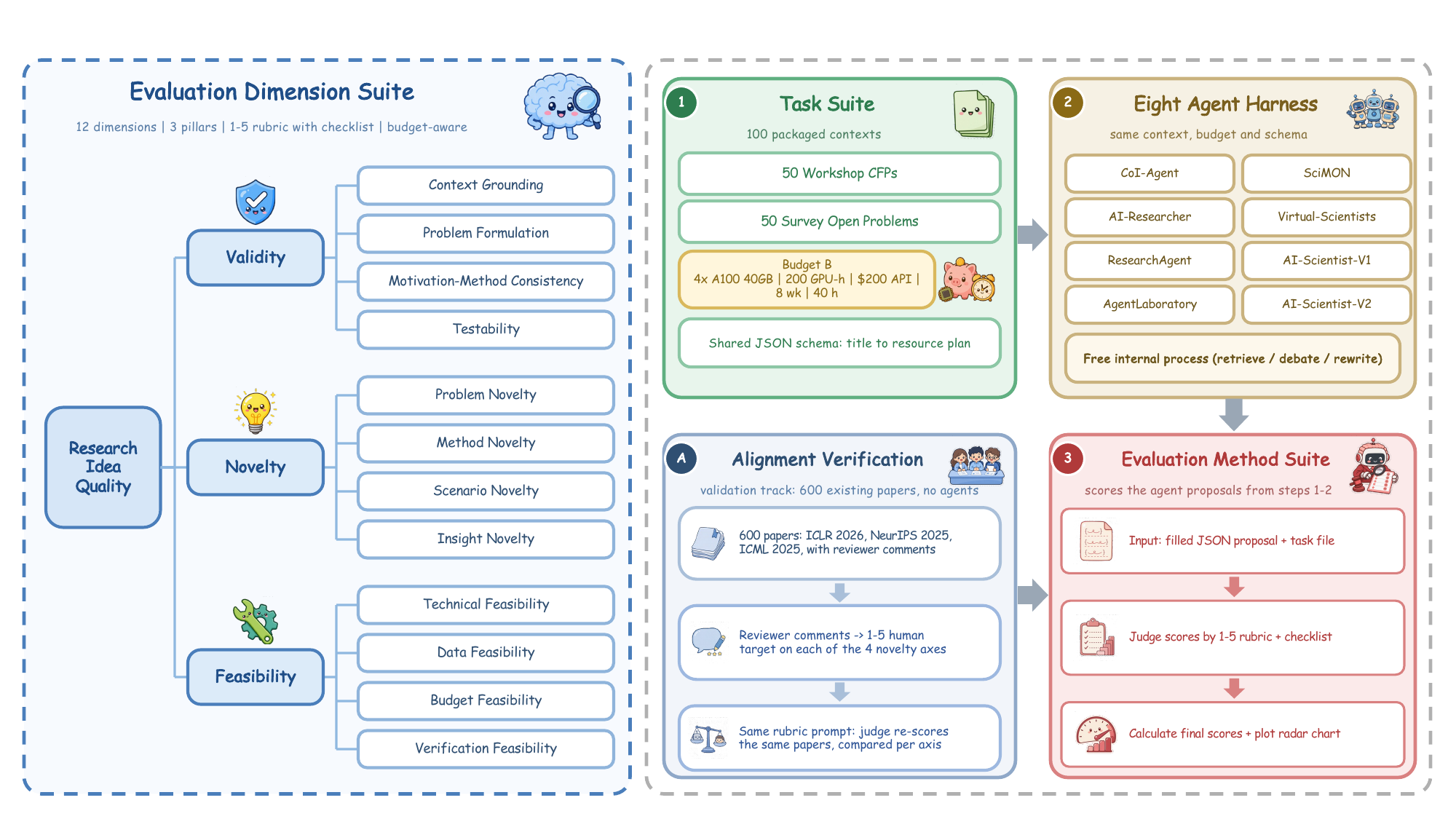}
% \end{center}
\vspace{-15pt}
\caption{Overview of AlphaIdeaBench. Alignment Verification maps reviewer comments on 600 papers onto the four novelty axes before those axes are used on agent writes.}
\label{fig:pipeline}
\vspace{-10pt}
\end{figure}

AlphaIdeaBench scores one structured proposal at a time. Figure~\ref{fig:pipeline} shows the pieces of that evaluation: a task file, the JSON the harness must return, a public 12-dimension rubric, a resource envelope, a judge, and an alignment check of the novelty axes against reviewer text. This section defines each piece in detail. Every component is held strictly identical across agents so differences reflect ideation quality rather than prompt engineering. The agreement numbers on the 600 papers, along with the empirical evaluation scores of the agent writes, are reported in Section~\ref{sec:experiments} for completeness and comparison.

\subsection{Benchmark construction}
\label{sec:task_construction}
% Each item is a scholarly context, a required JSON schema, and a resource envelope. Every harness sees the same three.

\paragraph{Task Collection.}
The suite has 100 tasks in two tracks of 50, matching the two ways a human researcher starts a project. The workshop track uses official calls hosted at NeurIPS, ICML, and ICLR from 2023 to 2025. Each item keeps the workshop narrative, the listed research questions, and the original topic scope, so the question is still partly open. The survey track uses 50 surveys. We take the open problems the authors themselves state, so the agent is asked for an intervention on a named bottleneck rather than for an open frontier. Every file carries the same academic envelope and the same required output, both defined next, so no harness gets a friendlier prompt.

The 100 tasks cover nine fields (Table~\ref{tab:task_distribution}): LLMs and reasoning (22), AI for science (13), efficiency and systems (13), multimodal and vision (12), trustworthy ML and safety (12), robotics and reinforcement learning (8), interpretability and theory (8), graphs and geometry (6), and continual learning (6). Figure~\ref{fig:domains}(b) and Figure~\ref{fig:domains}(c) show the same field mix and the keyword cloud. Appendix~\ref{app:task_examples} prints one file from each track (\emph{ws\_02} and \emph{survey\_19}).

\textbf{Required Output.}
A harness may retrieve, debate, or rewrite internally. It must return one JSON object. The schema requires a title, question, hypothesis, method, experiments, baselines, success criteria, related work, limitations, and a resource plan. These fields are what the judge reads. A missing baseline, a method that does not attack the stated question, or a resource line that ignores the envelope is then a property of the proposal. We do not score the internal trace. Role play and rewrite loops affect the result only through the JSON they emit, which is what makes harnesses with different control flow comparable across diverse agent architectures and search strategies.

\textbf{Resource Envelope.}
\label{subsec:budget_conditioning}
Feasibility is scored against a stated budget, not against an implicit guess that a lab could probably run the plan. Agents receive
$
\mathcal{B} = \langle \mathcal{C}_{\mathrm{gpu}},\; \mathcal{C}_{\mathrm{api}},\; \mathcal{T}_{\mathrm{cal}},\; \mathcal{H}_{\mathrm{human}} \rangle.
$
In the 100-task study every topic uses the same academic setting: four A100 40GB GPUs, 200 GPU hours, \$200 of API credit, eight calendar weeks, and 40 annotation hours. The judge compares the proposed compute and labor plan with this $\mathcal{B}$, and the agent sees the same numbers. Section~\ref{sec:experiments} later keeps the topic fixed and replaces the whole envelope. The student setting is one A100, 20 GPU hours, \$20 of API credit, two weeks, and 8 annotation hours. The lab setting is eight A100 GPUs, 2000 GPU hours, \$2000 of API credit, 16 weeks, and 80 annotation hours. A plan that becomes more ambitious only by ignoring the cap fails budget feasibility.

\subsection{Evaluation System}
\label{subsec:eval_system}
% The evaluation system scores that JSON on twelve dimensions, then turns the scores into harness means and a novelty check against reviewer text.

\textbf{Scoring Dimensions.}
\label{subsec:taxonomy}
The rubric is public, and the scoring rules were written by human experts. Each of the twelve checks is scored from 1 to 5, and Appendix~\ref{app:dimensions} releases the descriptor and the questions the judge must answer. A single overall score hides which part of the idea failed, and overlap with one published paper punishes a legitimate departure without saying whether the plan can be run. The twelve checks fall into three groups that do not substitute for one another: a proposal can be well grounded and still be a dataset swap, or novel and still exceed the envelope.
\begin{itemize}[leftmargin=*, topsep=0pt, partopsep=0pt, parsep=0pt, itemsep=0pt]
\item \textit{Validity.}
Context grounding penalizes hallucinated literature and phantom baselines. Problem formulation asks whether the topic is turned into a bounded, answerable question rather than a restatement of the prompt. Motivation and method consistency checks that the proposed method attacks the stated bottleneck. Testability requires observable outcomes, controls, and baselines that could falsify the stated claim rather than a success criterion that cannot fail once the plan is run. Together, these four checks establish empirical soundness: an idea must be anchored in real prior art and open to experimental falsification before it is judged on novelty.
\item \textit{Novelty.}
Problem novelty concerns the question or failure mode. Method novelty concerns a non-trivial change to the algorithm or the objective, not a rename of a known gadget. Scenario novelty concerns a new domain, distribution, or operating constraint that forces a redesign rather than a dataset swap. Insight novelty concerns a mechanism, causal factor, or boundary condition that changes how the problem is understood. A high method score and a high scenario score are different claims. These four axes are the ones Alignment Verification checks against reviewer text.
\item \textit{Feasibility.}
Technical feasibility rejects oracle information and circular dependencies, such as a method that needs the label it claims to predict. Data feasibility checks access, licenses, and labels. Budget feasibility compares the full experimental program, including baselines and ablations, with the envelope. The judge can use internet tools to look up API and GPU prices rather than trusting the agent's own arithmetic. Verification feasibility asks whether the metrics and the sample size can isolate the claimed effect rather than a metric that moves with every confound in the data.
\end{itemize}

Together, these twelve checks decouple scientific merit into orthogonal axes: a proposal cannot compensate for hallucinated grounding with algorithmic novelty, nor can an ambitious formulation bypass physical resource bounds. Evaluating them independently prevents aggregate metric compression and isolates precisely where automated ideation succeeds or fails across distinct dimensions of scientific proposal quality.

% \noindent

\textbf{Judge.}
The production judge is Gemini 3.7 Flash. It reads the JSON and the task file, and applies the 1 to 5 descriptors in Appendix~\ref{app:dimensions}. It does not see reviewer text, author identities, or accept decisions. Validity and feasibility have no human reference in this study: they are checks against the declared experiment and against $\mathcal{B}$. A second judge, GPT-5 mini, does not produce the main scores. It independently re-scores the same finished writes, which allows us to verify whether the relative agent ranking remains stable under an alternative judge rather than depending on a single scoring model (Appendix~\ref{app:judge_agreement}).

\textbf{Alignment Verification.}
Alignment Verification, the block in Figure~\ref{fig:pipeline}, checks the four novelty axes against real reviewer comments downloaded for these papers. Reviewers did not fill in a 1 to 5 score for problem, method, scenario, or insight. Before those axes are used on agent writes, Gemini 3.7 Flash reads those comments and maps each discussed facet to a ground-truth score from 1 to 5, following {NovBench} \citep{wu2026novbench}. The set is 600 papers, 200 each from ICLR 2026, NeurIPS 2025, and ICML 2025, stratified to each venue's official mean-score mix (Table~\ref{tab:sampling_distribution}). Facets that reviewers did not discuss are left missing, so the number of labeled items differs by axis. The comparison score is a separate pass: that judge sees the title and abstract only, not the reviews, so the experimental section is invisible both to the ground truth and to the judge.

\textbf{Metrics.}
\label{subsec:metrics}
Let \(s_{t,d}\in\{1,\ldots,5\}\) be the judge score of proposal \(t\) on dimension \(d\in\{1,\ldots,12\}\). With \(T\) proposals for a harness (\(T{=}100\) in the main tables),
\begin{align}
\mathrm{Overall}_t&=\tfrac{1}{12}\sum\nolimits_{d=1}^{12}s_{t,d},&
\bar{s}_d&=\tfrac{1}{T}\sum\nolimits_{t=1}^{T}s_{t,d},&
\mathrm{Overall}&=\tfrac{1}{12}\sum\nolimits_{d=1}^{12}\bar{s}_d.
\label{eq:overall}
\end{align}
Overall is an unweighted mean. A gain on one novelty axis is not rescaled by the eight validity and feasibility scores, but neither can a single novelty spike hide a plan that fails testability. We report the twelve means as well as Overall, because the mean alone compresses when validity and feasibility sit near the ceiling.
For one novelty axis, let \(\hat{y}_i\) and \(y_i\) be the judge score and the reviewer-mapped score on item \(i=1,\ldots,N\), and let \(p,q\) be the corresponding histograms on \(\{1,\ldots,5\}\).
\begin{equation}
\label{eq:novelty_metrics}
\resizebox{0.92\linewidth}{!}{$\displaystyle
\mathrm{DistAcc}=1-\frac{1}{2}\sum_{k=1}^{5}|p_k-q_k|,\quad
\text{Within }\pm 1=\frac{1}{N}\sum_{i=1}^{N}\mathbf{1}\bigl[|\hat{y}_i-y_i|\le 1\bigr],\quad
\text{Exact Acc}=\frac{1}{N}\sum_{i=1}^{N}\mathbf{1}[\hat{y}_i=y_i].
$}
\end{equation}
DistAcc is taken from {NovBench}~\citep{wu2026novbench}. It is one minus half the L1 distance between the two score histograms, so it checks whether the judge and the reviewers use the scale in the same way. Within $\pm 1$ is the fraction of items on which the two scores differ by at most one point, which is the agreement we treat as usable on a five-point rubric. Exact Acc is the fraction of ties. MAD is $\frac{1}{N}\sum_{i=1}^{N}|\hat{y}_i-y_i|$. Judge mean J$\mu$ and human mean H$\mu$ are the two averages, and Gap is J$\mu$ minus H$\mu$. A negative Gap means the judge is stricter than the mapped reviews. DistAcc, Within $\pm 1$, and Exact Acc are reported as percentages.
On the 50 surveys we also record which listed open problem the proposal actually writes. A hit is an explicit ``challenge $k$'' or ``C$k$'' tag, otherwise keyword overlap with a listed problem. Multi-challenge means more than one hit: $\mathrm{Multi}=\frac{1}{50}\sum_{t=1}^{50}\mathbf{1}[n_t>1]$, with $n_t$ the number of hits on survey $t$. This is a property of the text, not a thirteenth judge score. It is how Section~\ref{sec:experiments} separates a model prior, which picks the challenge, from a harness policy, which decides whether to dig one bottleneck or to fuse several.

\section{Experimental Evaluation}
\label{sec:experiments}

\textbf{Setup.}
\label{subsec:setup}
We run eight published harnesses on all 100 tasks with Gemini 3.7 Flash as the shared ideation backbone. Each topic yields one structured idea, scored on the 12 dimensions. For {CoI-Agent} we also rerun ideation with GPT-5.4 mini, Qwen 3.7 Flash, DeepSeek-V4 Flash, GPT-OSS-120B, Tencent HY3, Grok 4.3, and Llama 4 Scout, keeping the judge unchanged (Table~\ref{tab:coi_backbone}). Table~\ref{tab:agent_setup} records only the mechanism that actually runs in this study. Several of these systems originally continue into code execution or manuscript writing. Here they stop at the JSON proposal for a controlled comparison.

\begin{table*}[t]
% \vspace{-10pt}
\caption{Eight harnesses with their corresponding mechanism.}
\label{tab:agent_setup}
\begin{center}
\footnotesize
\setlength{\tabcolsep}{4pt}
\renewcommand{\arraystretch}{0.9}
\begin{tabular}{@{}lp{11cm}@{}}
\toprule
Agent & Mechanism used in the study \\
\midrule
CoI-Agent & Literature chain construction plus critique of the draft idea \citep{li2025coi} \\
SciMON & Inspiration retrieval, novelty collision check, and iterative rewrite \citep{wang2024scimon} \\
AI-Researcher & Survey to idea to plan expansion with a falsification oriented protocol \citep{tang2025airesearcher} \\
Virtual-Scientists & Multi persona brainstorming and consensus selection \citep{su2025virtualscientists} \\
ResearchAgent & Iterative retrieval and refinement against a literature graph \citep{baek2025researchagent} \\
AI-Scientist-V1 & JSON draft, self reflection, and a Semantic Scholar novelty check \citep{lu2024ai} \\
AgentLaboratory & PhD  style role play over literature and experimental design \citep{schmidgall2025agentlaboratory} \\
AI-Scientist-V2 & Template free self reflection loops \citep{yamada2025ai} \\
\bottomrule
\end{tabular}
\end{center}
% \vspace{-10pt}
\end{table*}

\begin{figure}[t]
\begin{center}
\begin{minipage}[c]{0.325\linewidth}
\centering
\includegraphics[width=\linewidth]{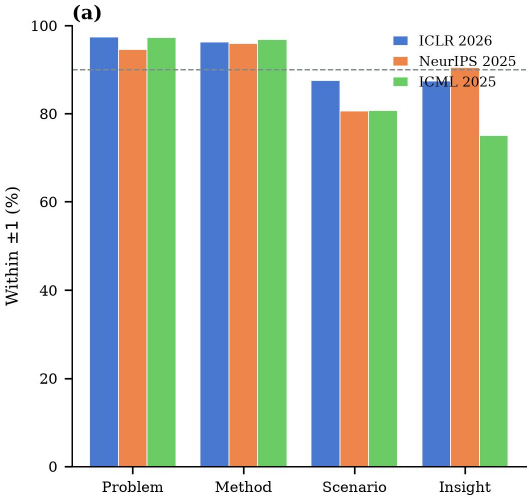}
\end{minipage}\hfill
\begin{minipage}[c]{0.325\linewidth}
\centering
\includegraphics[width=\linewidth]{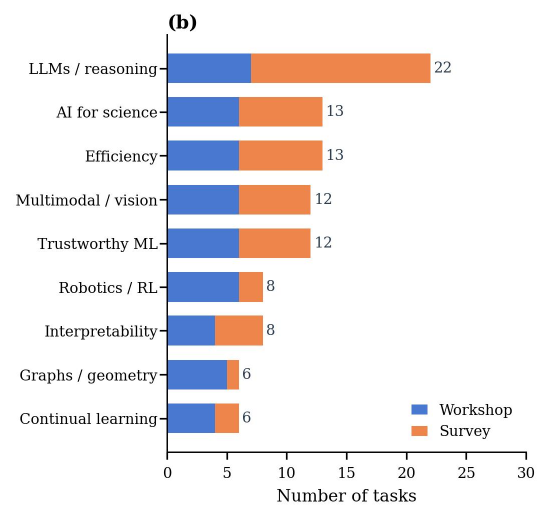}
\end{minipage}\hfill
\begin{minipage}[c]{0.325\linewidth}
\centering
\includegraphics[width=\linewidth]{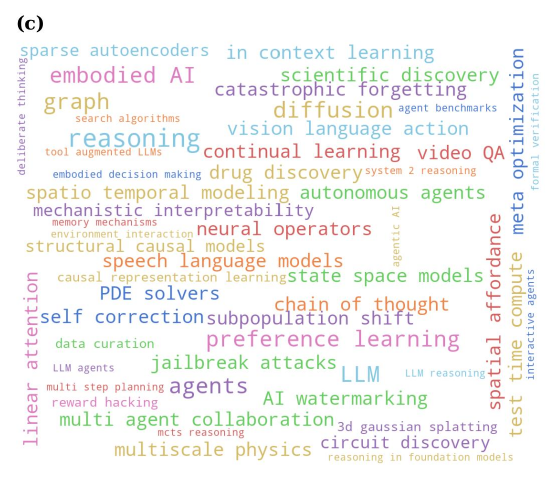}
\end{minipage}
\end{center}
\caption{(a) Within plus or minus one agreement of the  judge with reviewer mapped novelty. (b) Domain mix on the 100 task release. (c) Word cloud of task titles and curated keywords.}
\label{fig:meta_eval}
\label{fig:domains}
\end{figure}

\textbf{Novelty Alignment.}
\label{sec:meta_evaluation}
Table~\ref{tab:meta_eval_main} is the check we use before reading agent scores. For Gemini 3.7 Flash, problem and method novelty stay within one point of the reviewer-mapped score on 94\% to 97\% of items at every venue. DistAcc on problem is above 91\%, and insight DistAcc is 81.9\% to 83.8\%. Both sit above the best DistAcc in {NovBench}, 0.716 for SEA-S under zero-shot prompting \citep{wu2026novbench}. The Gaps on problem and method stay small. Scenario novelty is sparsely labeled ($N$ from 31 to 56). Within $\pm 1$ falls to about 80\%, and the Gap is negative ($-0.27$ to $-0.58$): the judge is stricter than reviewers (Figure~\ref{fig:meta_eval}(a)). The same table reports DeepSeek-V4-Flash and GPT-5 mini. Agreement is not uniform across judges or axes, which is why the 100-task ranking later uses one production judge and treats the second judge as a stability check rather than as ground truth.

\begin{table}[t]
% \vspace{-10pt}
\caption{Novelty axis alignment of three judges with reviewer-mapped scores across 600 papers (ICLR 2026, NeurIPS 2025, ICML 2025).}
\label{tab:meta_eval_main}
\begin{center}
\setlength{\tabcolsep}{2.0pt}
\renewcommand{\arraystretch}{0.90}
\resizebox{\linewidth}{!}{%
\begin{tabular}{@{}lrrrrrrrrrrrrrrrrrrrrrr@{}}
\toprule
 & & \multicolumn{7}{c}{Gemini 3.7 Flash} & \multicolumn{7}{c}{DeepSeek-V4-Flash} & \multicolumn{7}{c}{GPT-5 mini} \\
\cmidrule(lr){3-9}\cmidrule(lr){10-16}\cmidrule(lr){17-23}
Dimension & $N$ & DistAcc & $\pm 1$ & Exact & MAD & J$\mu$ & H$\mu$ & Gap & DistAcc & $\pm 1$ & Exact & MAD & J$\mu$ & H$\mu$ & Gap & DistAcc & $\pm 1$ & Exact & MAD & J$\mu$ & H$\mu$ & Gap \\
\midrule
\multicolumn{23}{@{}l}{\textbf{ICLR 2026} ($N{=}200$ papers)} \\
Problem & 114 & 91.2 & 97.4 & 52.6 & 0.50 & 3.02 & 2.97 & $+0.05$ & 64.4 & 86.7 & 39.3 & 0.74 & 2.98 & 3.01 & $-0.03$ & 76.3 & 87.4 & 38.5 & 0.76 & 3.21 & 3.01 & $+0.20$ \\
Method & 190 & 69.5 & 96.3 & 43.2 & 0.61 & 2.93 & 2.65 & $+0.28$ & 69.1 & 89.2 & 41.2 & 0.70 & 2.84 & 3.11 & $-0.27$ & 73.2 & 82.0 & 34.0 & 0.85 & 2.74 & 3.11 & $-0.38$ \\
Scenario & 56 & 75.0 & 87.5 & 39.3 & 0.73 & 2.41 & 2.68 & $-0.27$ & 79.1 & 84.9 & 39.5 & 0.77 & 2.98 & 2.74 & $+0.23$ & 74.4 & 83.7 & 46.5 & 0.70 & 3.14 & 2.74 & $+0.40$ \\
Insight & 111 & 82.9 & 87.4 & 41.4 & 0.73 & 2.91 & 2.83 & $+0.08$ & 74.4 & 84.8 & 34.4 & 0.83 & 2.97 & 2.82 & $+0.14$ & 84.8 & 82.4 & 39.2 & 0.84 & 3.09 & 2.82 & $+0.26$ \\
\midrule
\multicolumn{23}{@{}l}{\textbf{NeurIPS 2025} ($N{=}200$ papers)} \\
Problem & 109 & 93.6 & 94.5 & 53.2 & 0.52 & 3.05 & 2.96 & $+0.09$ & 90.8 & 97.2 & 56.9 & 0.46 & 3.06 & 2.96 & $+0.09$ & 89.0 & 95.4 & 51.4 & 0.53 & 3.20 & 2.96 & $+0.24$ \\
Method & 194 & 80.9 & 95.9 & 49.0 & 0.56 & 3.11 & 2.90 & $+0.21$ & 77.4 & 96.4 & 52.8 & 0.51 & 2.94 & 2.90 & $+0.04$ & 87.7 & 91.3 & 42.6 & 0.67 & 2.75 & 2.90 & $-0.14$ \\
Scenario & 36 & 61.1 & 80.6 & 38.9 & 0.81 & 2.44 & 2.86 & $-0.42$ & 80.6 & 97.2 & 66.7 & 0.36 & 3.00 & 2.86 & $+0.14$ & 91.7 & 100.0 & 72.2 & 0.28 & 2.97 & 2.86 & $+0.11$ \\
Insight & 105 & 81.9 & 90.5 & 49.5 & 0.62 & 3.19 & 2.97 & $+0.22$ & 83.8 & 91.4 & 45.7 & 0.64 & 3.19 & 2.97 & $+0.22$ & 84.8 & 90.5 & 53.3 & 0.58 & 3.23 & 2.97 & $+0.26$ \\
\midrule
\multicolumn{23}{@{}l}{\textbf{ICML 2025} ($N{=}200$ papers)} \\
Problem & 74 & 93.2 & 97.3 & 58.1 & 0.45 & 3.01 & 2.92 & $+0.09$ & 85.7 & 98.7 & 61.0 & 0.40 & 3.09 & 2.92 & $+0.17$ & 84.4 & 93.5 & 49.4 & 0.57 & 3.18 & 2.92 & $+0.26$ \\
Method & 190 & 81.6 & 96.8 & 47.9 & 0.55 & 3.14 & 3.03 & $+0.11$ & 73.2 & 96.9 & 46.4 & 0.57 & 2.98 & 3.04 & $-0.05$ & 80.9 & 91.2 & 38.7 & 0.70 & 2.71 & 3.04 & $-0.33$ \\
Scenario & 31 & 61.3 & 80.7 & 35.5 & 0.84 & 2.00 & 2.58 & $-0.58$ & 83.9 & 90.3 & 41.9 & 0.68 & 2.94 & 2.58 & $+0.35$ & 71.0 & 83.9 & 58.1 & 0.58 & 3.10 & 2.58 & $+0.52$ \\
Insight & 80 & 83.8 & 75.0 & 35.0 & 0.91 & 3.02 & 2.61 & $+0.41$ & 72.3 & 73.5 & 36.1 & 0.93 & 3.17 & 2.63 & $+0.54$ & 75.9 & 78.3 & 26.5 & 0.98 & 3.14 & 2.63 & $+0.52$ \\
\bottomrule
\end{tabular}%
}
\end{center}
\end{table}

\begin{figure}[t]
\begin{center}
\includegraphics[width=\linewidth]{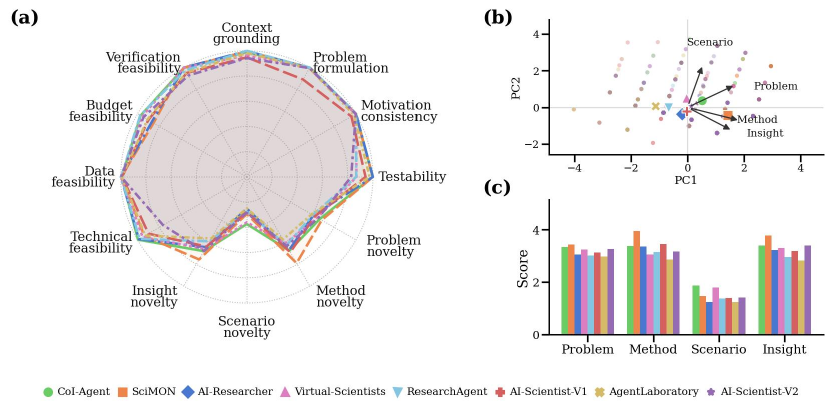}
\end{center}
\caption{(a) Twelve dimension radar of eight harnesses on 100 tasks.  (b) PCA of the z-scored four novelty axes, one point per task.  (c) Mean scores on the four novelty axes. }
\label{fig:radar_comparison}
\label{fig:novelty_scatter}
\label{fig:novelty_breakdown}
\vspace{-10pt}
\end{figure}

\begin{table}[t]
\caption{All twelve dimension means on 100 tasks for 8 harnesses (The backbone is Gemini 3.7 Flash.). Overall is the average of all dimensions. The high Overall band is not a claim that every model scores this high: Table~\ref{tab:coi_backbone} keeps the CoI harness and swaps only the ideation model, and Overall then runs from 4.31 down to 1.84.}
\label{tab:macro_performance}
\begin{center}
\scriptsize
\setlength{\tabcolsep}{2.2pt}
\renewcommand{\arraystretch}{0.90}
\resizebox{\linewidth}{!}{%
\begin{tabular}{@{}lcccccccc@{}}
\toprule
Dimension & CoI-Agent & SciMON & AI-Researcher & Virtual-Scientists & ResearchAgent & AI-Scientist-V1 & AgentLaboratory & AI-Scientist-V2 \\
\midrule
Context grounding & 4.95 & 4.98 & 4.99 & 4.82 & \textbf{5.00} & 4.74 & 4.91 & 4.71 \\
Problem formulation & \textbf{5.00} & \textbf{5.00} & 4.99 & 4.97 & \textbf{5.00} & 4.46 & 4.97 & 4.98 \\
Motivation consistency & \textbf{5.00} & 4.98 & \textbf{5.00} & \textbf{5.00} & \textbf{5.00} & 4.79 & 4.91 & 4.99 \\
Testability & 4.99 & \textbf{5.00} & \textbf{5.00} & 4.35 & 4.34 & 4.71 & 4.87 & 4.14 \\
Problem novelty & 3.34 & \textbf{3.43} & 3.05 & 3.24 & 3.02 & 3.12 & 2.97 & 3.27 \\
Method novelty & 3.38 & \textbf{3.95} & 3.35 & 3.05 & 3.14 & 3.45 & 2.87 & 3.17 \\
Scenario novelty & \textbf{1.87} & 1.47 & 1.25 & 1.80 & 1.38 & 1.39 & 1.25 & 1.42 \\
Insight novelty & 3.40 & \textbf{3.78} & 3.22 & 3.30 & 2.95 & 3.19 & 2.82 & 3.39 \\
Technical feasibility & \textbf{4.98} & 4.69 & 4.94 & 4.77 & 4.91 & 4.50 & 4.65 & 3.83 \\
Data feasibility & 4.96 & 4.98 & \textbf{4.99} & \textbf{4.99} & 4.98 & 4.94 & 4.95 & 4.98 \\
Budget feasibility & 4.86 & 4.27 & 4.35 & 4.79 & \textbf{4.88} & 4.51 & 4.55 & 4.73 \\
Verification feasibility & 4.99 & \textbf{5.00} & \textbf{5.00} & \textbf{5.00} & 4.88 & 4.76 & 4.78 & 4.64 \\
\midrule
Overall & \textbf{4.31} & 4.29 & 4.18 & 4.17 & 4.12 & 4.05 & 4.04 & 4.02 \\
\bottomrule
\end{tabular}%
}
\end{center}
\end{table}

\textbf{Twelve-dimension Scores.}
% On the eight harnesses that share Gemini 3.7 Flash, overall spans  4.02 to 4.31 (Table~\ref{tab:macro_performance}). Validity and feasibility already sit near the ceiling, so the rank is a novelty rank (Figure~\ref{fig:radar_comparison}). The four novelty axes do not move together. {SciMON} leads method (3.95) and insight (3.78). {CoI-Agent} leads scenario (1.87), and every harness stays at or below that.
On the eight harnesses that share Gemini 3.7 Flash, overall spans 4.02 to 4.31 (Table~\ref{tab:macro_performance}). Validity and feasibility already sit near the ceiling, so the rank is a novelty rank (Figure~\ref{fig:radar_comparison}). The four novelty axes do not move together. In the PCA projection (Figure~\ref{fig:radar_comparison}(b)), PC1 captures method and insight innovation, where {SciMON} leads method (3.95) and insight (3.78) through inspiration collision. PC2 captures scenario shift, where {CoI-Agent} leads scenario (1.87) by modifying operating environments. Novelty pursuit also exposes feasibility trade-offs: {SciMON} drops to 4.27 on budget feasibility due to sprawling ablations, while {AI-Scientist-V2} incurs a technical feasibility penalty (3.83) from ungrounded execution loops. In contrast, {CoI-Agent} combines high scenario novelty with strict budget discipline across the full task set.

\begin{figure}[t]
\begin{center}
\includegraphics[width=\linewidth]{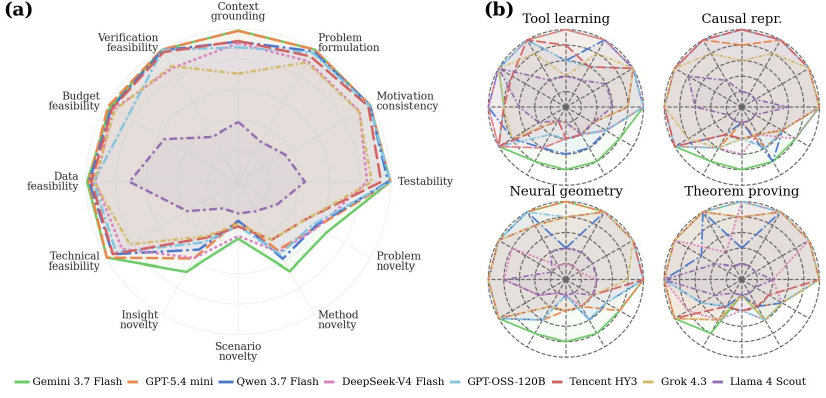}
\end{center}
\caption{Left: Twelve dimension means on the same 100 tasks. Right: the same eight ideation models on four topics, one idea each.}
\label{fig:coi_backbone}
\end{figure}

\textbf{Ideation Model.}
% Holding the CoI harness and the judge fixed, swapping only the ideation model drops overall from 4.31 with Gemini 3.7 Flash to 1.84 with Llama 4 Scout (Table~\ref{tab:coi_backbone}, Figure~\ref{fig:coi_backbone}). The loss is on validity, novelty, and feasibility together, so a weak model does not follow the literature or the envelope. A second judge does not undo the harness ranking: GPT-5 mini still puts {CoI-Agent} first (Spearman $\rho{=}0.81$; Appendix~\ref{app:judge_agreement}). On this CoI model ladder the top flips to GPT-5.4 mini ($\rho{=}0.83$), and Llama 4 Scout stays last.
Holding the CoI harness and the judge fixed, swapping only the ideation model drops overall from 4.31 with Gemini 3.7 Flash to 1.84 with Llama 4 Scout (Table~\ref{tab:coi_backbone}, Figure~\ref{fig:coi_backbone}). The loss spans validity, novelty, and feasibility together, showing that weak models fail literature grounding (1.95) and budget limits (2.80) rather than just novelty. A second judge does not undo the harness ranking: GPT-5 mini still puts {CoI-Agent} first (Spearman $\rho{=}0.81$; Appendix~\ref{app:judge_agreement}), while the top flips to GPT-5.4 mini ($\rho{=}0.83$) on the model ladder and Llama 4 Scout stays last.

\begin{table}[t]
\caption{All twelve dimension means for {CoI-Agent} on 100 tasks with eight ideation models.}
\label{tab:coi_backbone}
\begin{center}
\scriptsize
\setlength{\tabcolsep}{2.2pt}
\renewcommand{\arraystretch}{0.90}
\resizebox{\linewidth}{!}{%
\begin{tabular}{@{}lcccccccc@{}}
\toprule
Dimension & Gemini 3.7 Flash & GPT-5.4 mini & Qwen 3.7 Flash & DeepSeek-V4 Flash & GPT-OSS-120B & Tencent HY3 & Grok 4.3 & Llama 4 Scout \\
\midrule
Context grounding & \textbf{4.95} & 4.94 & 4.59 & 4.54 & 4.40 & 4.61 & 3.54 & 1.95 \\
Problem formulation & \textbf{5.00} & \textbf{5.00} & 4.93 & 4.56 & 4.88 & 4.74 & 4.51 & 1.51 \\
Motivation consistency & \textbf{5.00} & \textbf{5.00} & 4.99 & 4.60 & 4.97 & 4.93 & 4.58 & 1.78 \\
Testability & 4.99 & \textbf{5.00} & 4.92 & 4.23 & \textbf{5.00} & 4.69 & 4.39 & 2.19 \\
Problem novelty & \textbf{3.34} & 2.97 & 2.92 & 3.00 & 2.82 & 2.51 & 2.52 & 1.48 \\
Method novelty & \textbf{3.38} & 2.59 & 2.92 & 2.67 & 2.75 & 2.20 & 2.21 & 1.12 \\
Scenario novelty & \textbf{1.87} & 1.28 & 1.28 & 1.78 & 1.45 & 1.47 & 1.39 & 1.04 \\
Insight novelty & \textbf{3.40} & 2.91 & 2.57 & 2.89 & 2.30 & 2.12 & 2.06 & 1.01 \\
Technical feasibility & \textbf{4.98} & \textbf{4.98} & 4.76 & 4.41 & 4.63 & 4.75 & 4.11 & 1.93 \\
Data feasibility & \textbf{4.96} & 4.91 & 4.89 & 4.69 & 4.75 & 4.82 & 4.73 & 3.57 \\
Budget feasibility & 4.86 & \textbf{4.93} & 4.81 & 4.73 & 4.20 & 4.78 & 4.68 & 2.80 \\
Verification feasibility & 4.99 & \textbf{5.00} & 4.97 & 4.30 & 4.98 & 4.90 & 4.37 & 1.70 \\
\midrule
Overall & \textbf{4.31} & 4.13 & 4.05 & 3.87 & 3.93 & 3.88 & 3.59 & 1.84 \\
\bottomrule
\end{tabular}%
}
\end{center}
\end{table}

% \begin{wrapfigure}{r}{0.5\linewidth}
% \centering
% \includegraphics[width=\linewidth]{figure/alphaideabench.pdf}
% \caption{Eight writes on \emph{survey\_19}: C3 at $S{\le}2$, \textsc{CoI-Agent} at $S{=}4$ via C4.}
% \label{fig:casestudy_box}
% \end{wrapfigure}

\begin{figure}[t]
\begin{center}
\includegraphics[width=\linewidth]{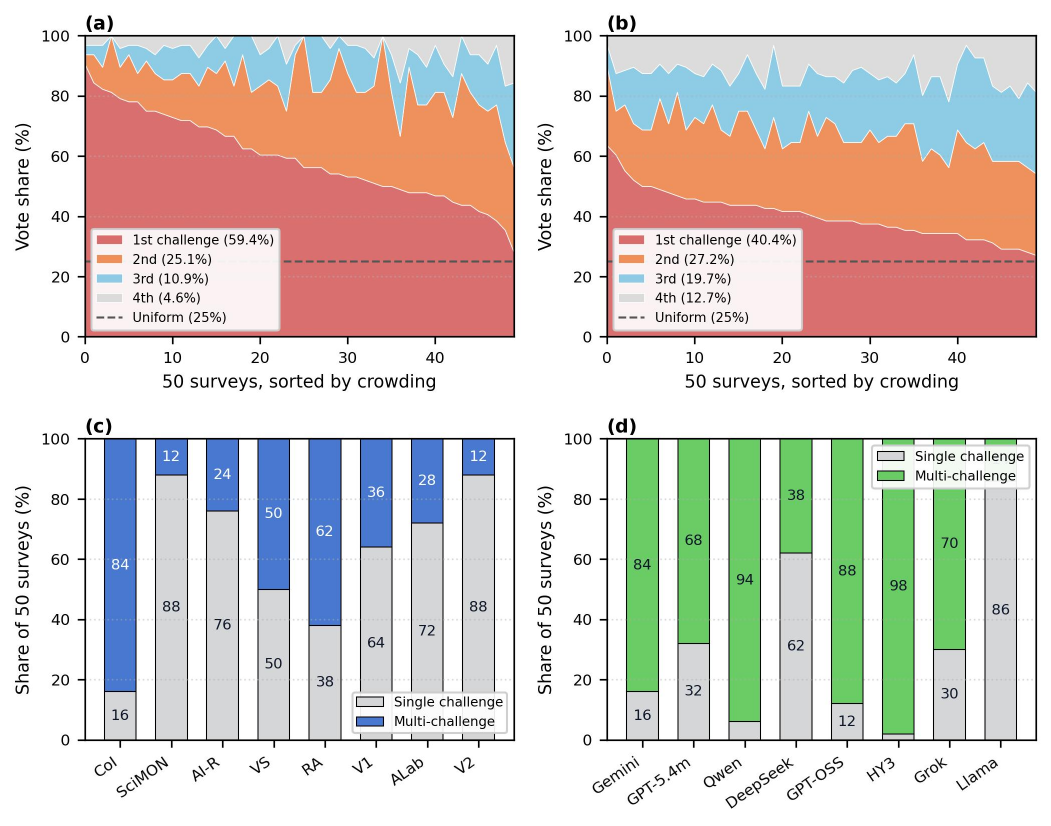}
\end{center}
\vspace{-10pt}
\caption{Challenge choice on the 50 surveys. (a)~Same Gemini, eight harnesses. (b)~Same {CoI-Agent}, eight models. (c)~Single versus fused writes, same Gemini. (d)~{CoI-Agent} with eight models.}
\label{fig:harness_vs_llm}
\vspace{-10pt}
\end{figure}

\textbf{Harness Versus Model.}
\label{subsec:harness_vs_llm}
Which listed challenge is written is mostly an LLM prior. On the 50 surveys, eight harnesses that share Gemini 3.7 Flash still pile 59.4\% of votes on the locally majority challenge and only 4.6\% on the fourth (Figure~\ref{fig:harness_vs_llm}(a)). Changing the scaffold barely moves that crowding. Holding {CoI-Agent} fixed and swapping only the ideation model spreads the mass: the majority share falls to 40.4\% and the fourth challenge rises to 12.7\% (Figure~\ref{fig:harness_vs_llm}(b)). How the idea is assembled is the harness. With the same Gemini backbone, {CoI-Agent} fuses more than one listed challenge on 84\% of surveys, while {SciMON} and {AI-Scientist-V2} stay on a single challenge on 88\% (Figure~\ref{fig:harness_vs_llm}(c)). We then keep that multi-challenge CoI scaffold and swap the model (Figure~\ref{fig:harness_vs_llm}(d)). Most backbones still fuse (Tencent HY3, Qwen 3.7 Flash, and GPT-OSS-120B at 88\% to 98\%). Llama 4 Scout does not execute the policy: it collapses to one challenge on 86\% of surveys, despite receiving the same explicit multi-challenge scaffold.

\begin{wrapfigure}{r}{0.5\linewidth}
\centering
\includegraphics[width=\linewidth]{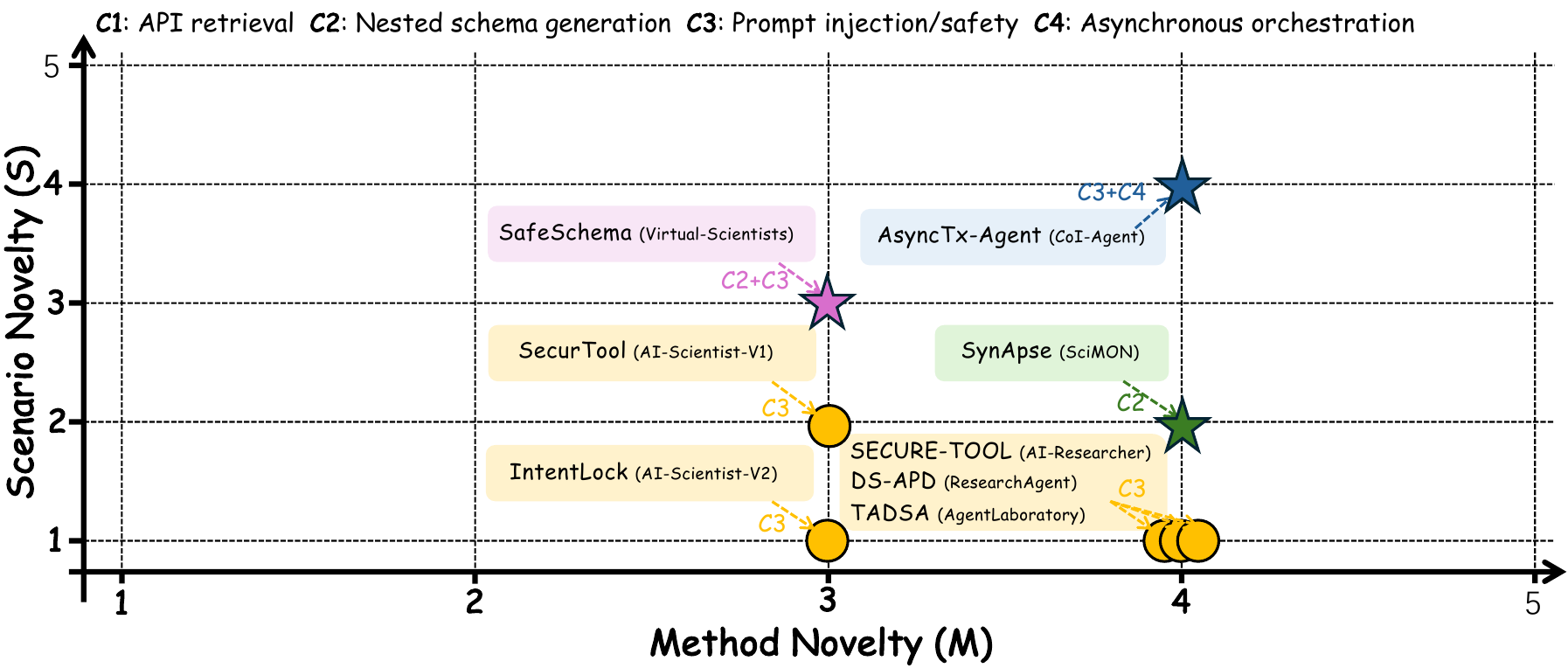}
\caption{Eight writes on \emph{survey\_19}: C3 at $S{\le}2$, \textsc{CoI-Agent} at $S{=}4$ via C4.}
\label{fig:casestudy_box}
\end{wrapfigure}

\textbf{Workshop Versus Survey.}
\label{subsec:modality_sensitivity}
Pooled workshop and survey means almost coincide (at most 0.17; Figure~\ref{fig:modality_radar}). Figure~\ref{fig:casestudy_radar} plots the same eight harnesses on seven topics. The panels do not share a winner, so a modality average is the wrong place to read who leaves the cluster.

\textbf{Budget.}
\label{subsec:budget_ablation}
The 100-task study holds $\mathcal{B}$ at the academic setting. To test whether a harness writes a different plan when the numbers change, we keep {CoI-Agent} and the Gemini 3.7 Flash judge and replace the whole budget on three surveys that differ in cost: tool use, vision-language-action, and mechanistic interpretability. The agent and the judge see the same $\mathcal{B}$, making the resource constraint explicit to both sides.

\begin{figure}[H]
\begin{center}
\includegraphics[width=\linewidth]{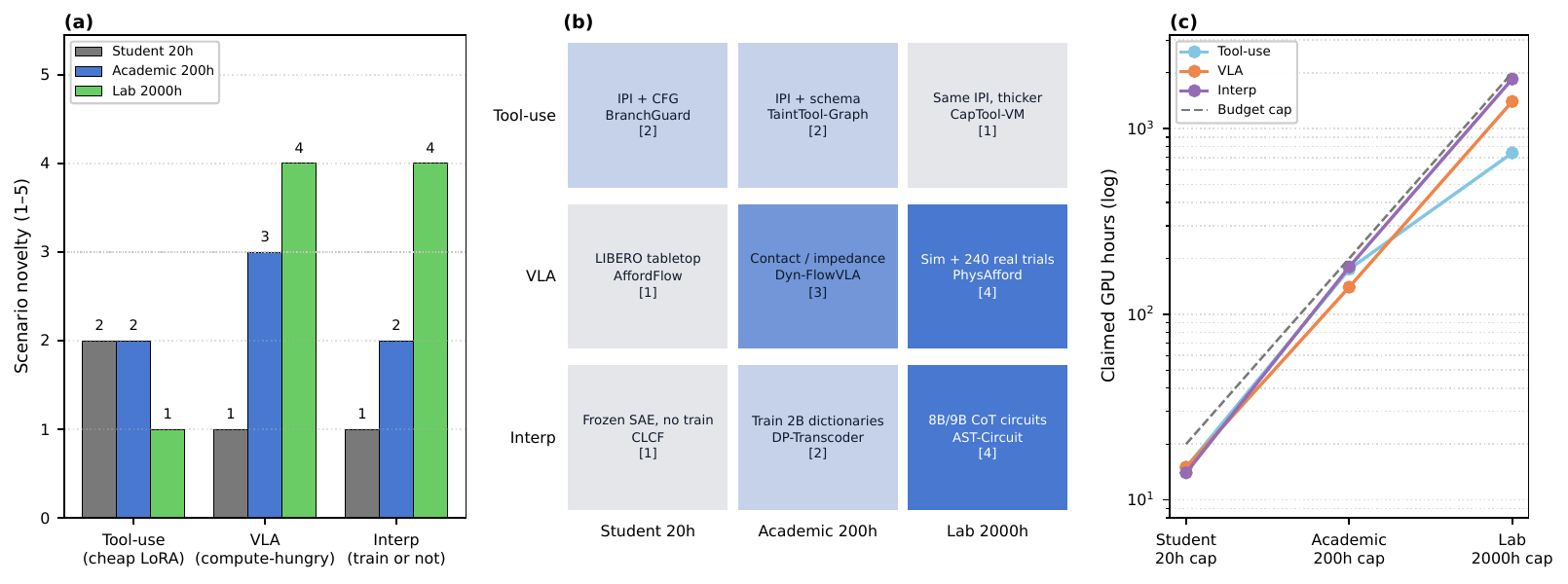}
\end{center}
\caption{{CoI-Agent} under three budgets. (a)~Scenario novelty. (b)~What the write is about. (c)~Claimed GPU hours versus the cap.}
\label{fig:budget_ablation}
\end{figure}

The harness does write different plans. Every cell stays inside its own cap (Figure~\ref{fig:budget_ablation}(c)), and the nine writes are not one template with a rescaled GPU line. Extra budget buys a more novel idea only on compute-heavy topics. On vision-language-action the operating world moves from a LIBERO tabletop in simulation, to contact and impedance, to simulation plus 240 real-robot trials once the lab budget is large enough. On interpretability the trained object scales: frozen sparse autoencoders with no training, then 2B dictionaries, then 8B and 9B chain-of-thought circuits. On a cheap and already instrumented topic the extra money does not change the research world. Tool use stays in prompt injection plus public benchmarks. The three writes differ in how many tests they run. At the lab cap, tool use claims only 740 of 2000 GPU hours, while vision-language-action and interpretability spend 1400 and 1850. A larger budget alters the experimental world only when compute is the binding constraint, and otherwise accumulates as additional tests of the same baseline setting rather than forcing novelty where resources are already sufficient.

\FloatBarrier

% \section{Limitations}
% \label{sec:Limitations}

\section{Limitations and Conclusion}
\label{sec:Conclusion}

% \textbf{Limitations.} The 100-task study is a controlled comparison, not a census of scientific ideation. Fifty workshop calls and fifty surveys all come from machine learning: ICLR-style CFPs and ML surveys. Medicine, biology, chemistry, and other wet-lab fields are absent, so crowding toward a majority challenge, scenario ceilings, and budget sensitivity may look different where public benches are thinner or experiments cannot be simulated. We compare eight published harnesses and, under {CoI-Agent}, eight ideation models. That is enough to separate a backbone prior from a search policy, but it is not a full model zoo, and new agents or later LLMs can move the Table~\ref{tab:macro_performance} and Table~\ref{fig:coi_backbone} order.

\textbf{Limitations.} The 100 tasks are drawn from machine learning, so findings on challenge crowding and scenario ceilings may behave differently in wet lab fields where experiments cannot be simulated. Moreover, while our comparison separates backbone priors from search policies across eight harnesses, it is not an exhaustive model zoo, and newer models may alter the ranking as model capabilities continue to evolve.

\textbf{Conclusion.} AlphaIdeaBench scores structured research proposals under a shared academic budget and a 12-dimension rubric, and it checks the novelty axes against reviewer text from 600 ML conference papers. We evaluated eight types of harnesses and several ideation models within this framework; CoI-Agent performed relatively well across the various metrics. Subsequently, during the process of decoupling the LLM from the harness, we found the following: the backbone largely determines which listed challenge looks worth writing, and the harness decides whether to dig one bottleneck or to fuse several. However, less capable LLMs cannot follow the harness instructions or the stated budget well. Extra budget buys a more ambitious write on compute-heavy topics and is absorbed as extra tests on cheap ones rather than opening a new research world on those topics. With a strong backbone, validity and feasibility sit near the ceiling, so harness rank is mostly about novelty. Scenario novelty stays low for every system we ran. We release the tasks, the rubric, and the reviewer-mapped scores so later agents can be compared under the same envelope and the same judge. Later work can keep this evaluation protocol fixed and change only the agent under test as automated research systems continue to improve.

% \section*{Ethics Statement}

% This study does not involve human subjects or sensitive personal data. All workshop calls and peer reviews are sourced from public repositories with reviewer identities withheld. All prompts and evaluation artifacts are released to ensure transparent and reproducible research.

\section*{Reproducibility Statement}

All 100 task files, scoring rubrics, and evaluation pipelines are open-sourced to ensure full empirical reproducibility. The complete codebase, benchmark datasets, and execution scripts are available at https://github.com/SHUNCHAOZHOU/AlphaIdeaBench.

\clearpage
\bibliography{iclr2027_conference}
\bibliographystyle{iclr2027_conference}

\appendix
% \section{Code and dataset}
% The code and dataset are available: https://anonymous.4open.science/r/AlphaIdeaBench-3122.
\section{Meta evaluation and domain tables}
\label{app:meta_tables}

\begin{figure*}[t]
\begin{center}
\includegraphics[width=\textwidth]{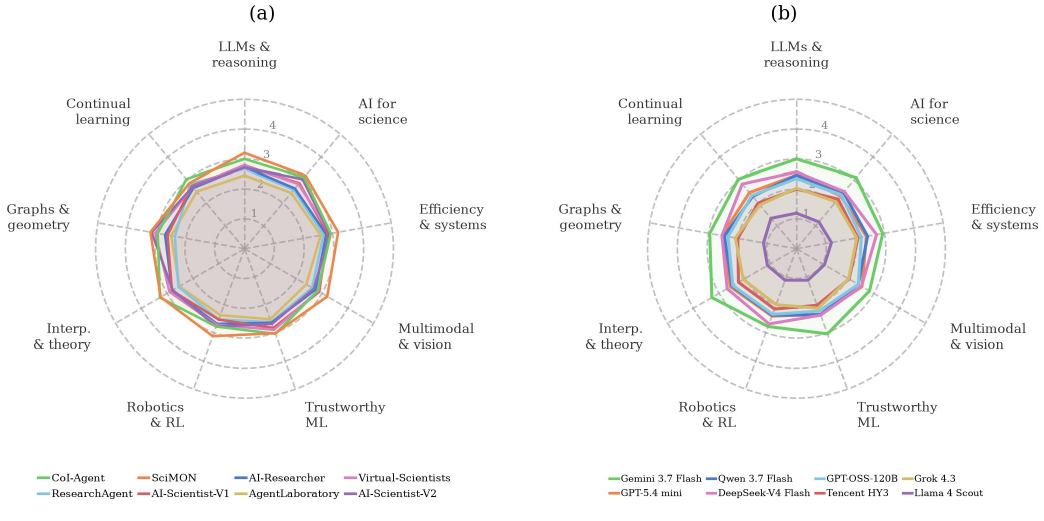}
\end{center}
\caption{(a) Mean novelty on the nine field bins of Figure~\ref{fig:domains}(b), eight harnesses.
(b) The same bins, {CoI-Agent} only, eight ideation models. The judge is Gemini 3.7 Flash in every trace. 
}
\label{fig:domain_radar}
\end{figure*}

Figure~\ref{fig:domain_radar} uses the same nine field bins as Figure~\ref{fig:domains}(b). Each spoke is mean novelty for that field, pooled over the topics in the bin. Domain means only span 0.18, so the field tag does not rewrite the aggregate ranking. The harness shapes do move a little: {SciMON} is the outer ring on six of nine bins and dips on continual learning (2.88). {CoI-Agent} is outer on the other three, interpretability and theory (3.28), trustworthy ML, and continual learning. {AI-Scientist-V2} is close to {SciMON} on graphs and geometry (3.17). {AgentLaboratory} is the inner ring on eight bins; {ResearchAgent} is lowest on graphs. Panel (b) keeps that binning and swaps only the CoI ideation model. Gemini 3.7 Flash is the outer ring on every bin and still peaks on interpretability and theory (3.28). DeepSeek-V4 Flash, GPT-5.4 mini, Qwen 3.7 Flash, and GPT-OSS-120B sit inside 2.21 to 2.83. Tencent HY3 and Grok 4.3 span 1.92 to 2.25 across bins, and Llama 4 Scout is in the innermost ring across all nine domains (1.08 to 1.33).

\begin{table}[h]
\caption{Proportional stratified sample of 600 papers (200 per venue). Pool is the official mean-score mix; sample is the Hamilton draw.}
\label{tab:sampling_distribution}
\begin{center}
\small
\setlength{\tabcolsep}{8pt}
\renewcommand{\arraystretch}{1.18}
\begin{tabular}{llcccc}
\toprule
Venue & Split & Low & \shortstack{Borderline\\low} & \shortstack{Borderline\\high} & High \\
\midrule
ICLR 2026 & Pool & 32.3\% & 50.1\% & 13.8\% & 3.8\% \\
 & Sample & 31.5\% & 51.0\% & 14.0\% & 3.5\% \\
\midrule
NeurIPS 2025 & Pool & 10.0\% & 50.2\% & 32.8\% & 7.0\% \\
 & Sample & 10.0\% & 50.0\% & 33.0\% & 7.0\% \\
\midrule
ICML 2025 & Pool & 15.9\% & 42.3\% & 26.0\% & 15.8\% \\
 & Sample & 16.0\% & 42.5\% & 25.5\% & 16.0\% \\
\bottomrule
\end{tabular}
\end{center}
{\footnotesize Pool sizes: ICLR $13{,}119$, NeurIPS $5{,}540$, ICML $3{,}256$. Sample $N{=}200$ (counts: ICLR $63/102/28/7$; NeurIPS $20/100/66/14$; ICML $32/85/51/32$). Bands are mean reviewer score. ICLR ($1$--$10$): ${<}4.0$, $[4.0,5.5)$, $[5.5,6.5)$, ${\ge}6.5$. NeurIPS ($1$--$6$): ${<}4.0$, $[4.0,4.5)$, $[4.5,5.0)$, ${\ge}5.0$. ICML ($1$--$5$): ${<}3.0$, $[3.0,3.3)$, $[3.3,3.7)$, ${\ge}3.7$.}
\end{table}

\begin{table}[h]
\caption{Domain counts on the 100 task Workshop and Survey release.}
\label{tab:task_distribution}
\begin{center}
\small
\begin{tabular}{lccc}
\toprule
Domain & Workshop & Survey & Total \\
\midrule
LLMs and reasoning & 7 & 15 & 22 \\
AI for science & 6 & 7 & 13 \\
Efficiency and systems & 6 & 7 & 13 \\
Multimodal and vision & 6 & 6 & 12 \\
Trustworthy ML and safety & 6 & 6 & 12 \\
Robotics and RL & 6 & 2 & 8 \\
Interpretability and theory & 4 & 4 & 8 \\
Graphs and geometry & 5 & 1 & 6 \\
Continual learning and adaptation & 4 & 2 & 6 \\
\midrule
Total & 50 & 50 & 100 \\
\bottomrule
\end{tabular}
\end{center}
\end{table}

\clearpage

\begin{figure}[t]
\begin{center}
\includegraphics[width=0.8\linewidth]{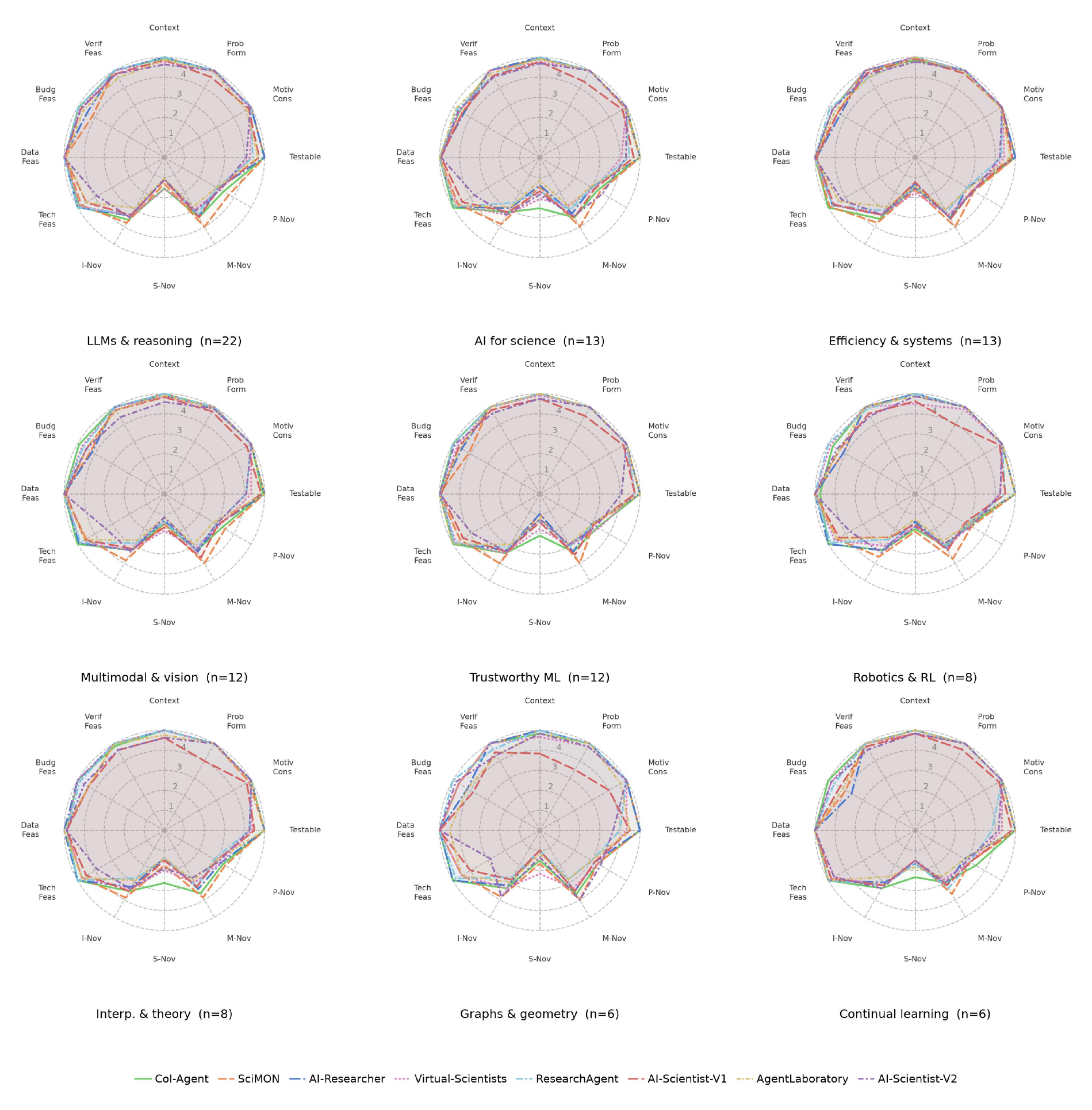}
\end{center}
\caption{Per domain 12 dimension radars on the nine field bins of Figure~\ref{fig:domains}(b). Each panel is one field and eight harnesses. 
% The family that fills the notch is not the same in every panel.
}
\label{fig:domain_radar_grid}
\end{figure}

Figure~\ref{fig:domain_radar_grid} keeps the twelve dimension layout and splits the 100 tasks by the same field tags. The validity ceiling and the novelty notch appear in every panel, and the notch is scenario. {SciMON} is farther out on method and insight in most bins, even the three where Figure~\ref{fig:domain_radar}(a) gives {CoI-Agent} the higher novelty mean: there {CoI-Agent} is farther out on scenario novelty, 2.62, 2.08, and 2.33 against {SciMON} 1.75, 1.00, and 1.50 on interpretability, trustworthy ML, and continual learning. {AI-Scientist-V2} meets {SciMON} on method and insight for graphs and geometry.

\begin{figure}[t]
\begin{center}
\includegraphics[width=\linewidth]{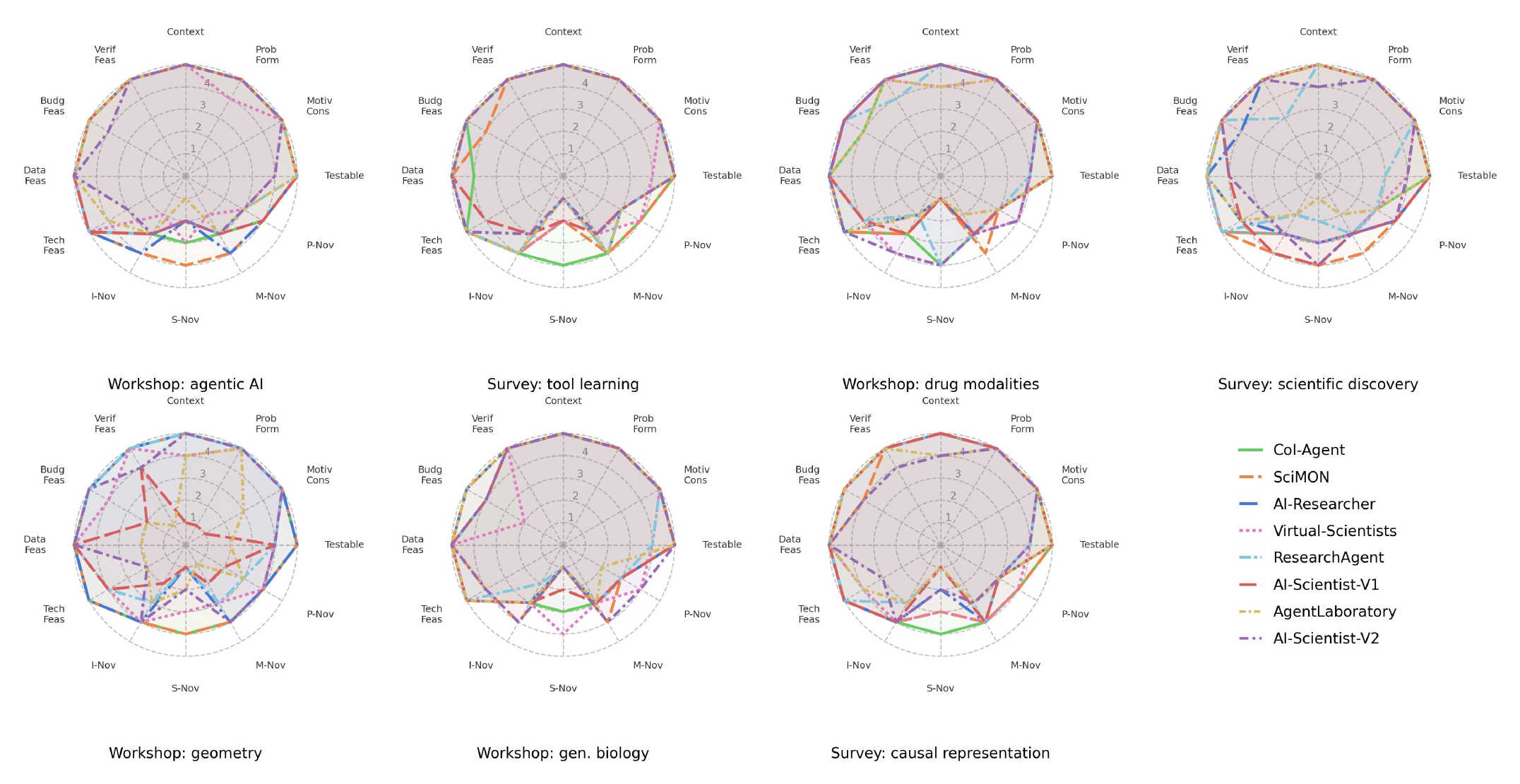}
\end{center}
\caption{Per topic 12 dimension radars on seven tasks. Score tables and excerpts are in Appendix~\ref{app:casestudy}.}
\label{fig:casestudy_radar}
\end{figure}

\clearpage
\section{Workshop and survey task files}
\label{app:task_examples}

The tracks differ only in the scholarly context. 

\textbf{Shared Budget and Schema.}
Four A100 40GB GPUs, 200 GPU hours, \$200 of API credit, eight calendar weeks, and 40 annotation hours. Required JSON fields: title, question, hypothesis, method, experiments, baselines, success criteria, related work, limitations, resource plan.

\begin{center}
\fbox{%
\begin{minipage}{0.94\linewidth}
\scriptsize
\noindent\textbf{Workshop example.} \emph{ws\_02\_agentic\_ai}. ICLR 2025 CFP.\\
\textit{Title.} Towards Agentic AI for Science: Hypothesis Generation, Comprehension, Quantification, and Validation.\\
\textit{Context (abridged).} The CFP asks for fully autonomous scientific agents and lists four open thrusts. It does not name one bottleneck.
\begin{itemize}
\setlength{\itemsep}{0pt}
\setlength{\parsep}{0pt}
\item Thrust 1: design of agentic systems (scientific tools, multi-agent decomposition, human in the loop).
\item Thrust 2: theory (quantification, validation, guarantees).
\item Thrust 3: application (domain adaptation, trust, ethics).
\item Thrust 4: open problems (knowledge curation, continual update, reproducibility).
\end{itemize}
\end{minipage}}
\end{center}

\begin{center}
\fbox{%
\begin{minipage}{0.94\linewidth}
\scriptsize
\noindent\textbf{Survey example.} \emph{survey\_19\_agentic\_tool\_use}.\\
\textit{Title.} Tool Learning with Foundation Models.\\
\textit{Context (abridged).} Author-stated open problems, numbered as challenges. The write must target at least one.
\begin{itemize}
\setlength{\itemsep}{0pt}
\setlength{\parsep}{0pt}
\item C1: retrieval and selection among thousands of APIs when documentation exceeds the context window.
\item C2: nested schema and parameter compliance at decode time.
\item C3: indirect prompt injection from untrusted tool outputs.
\item C4: concurrent asynchronous orchestration of tool calls.
\end{itemize}
\end{minipage}}
\end{center}

% \clearpage
\section{Scoring rubrics and 12 way scores}
\label{app:dimensions}

\input{scoring_rubrics.tex}

\clearpage
\section{Workshop versus survey radars}
\label{app:modality_grid}
\begin{figure}[h]
\begin{center}
\includegraphics[width=\linewidth]{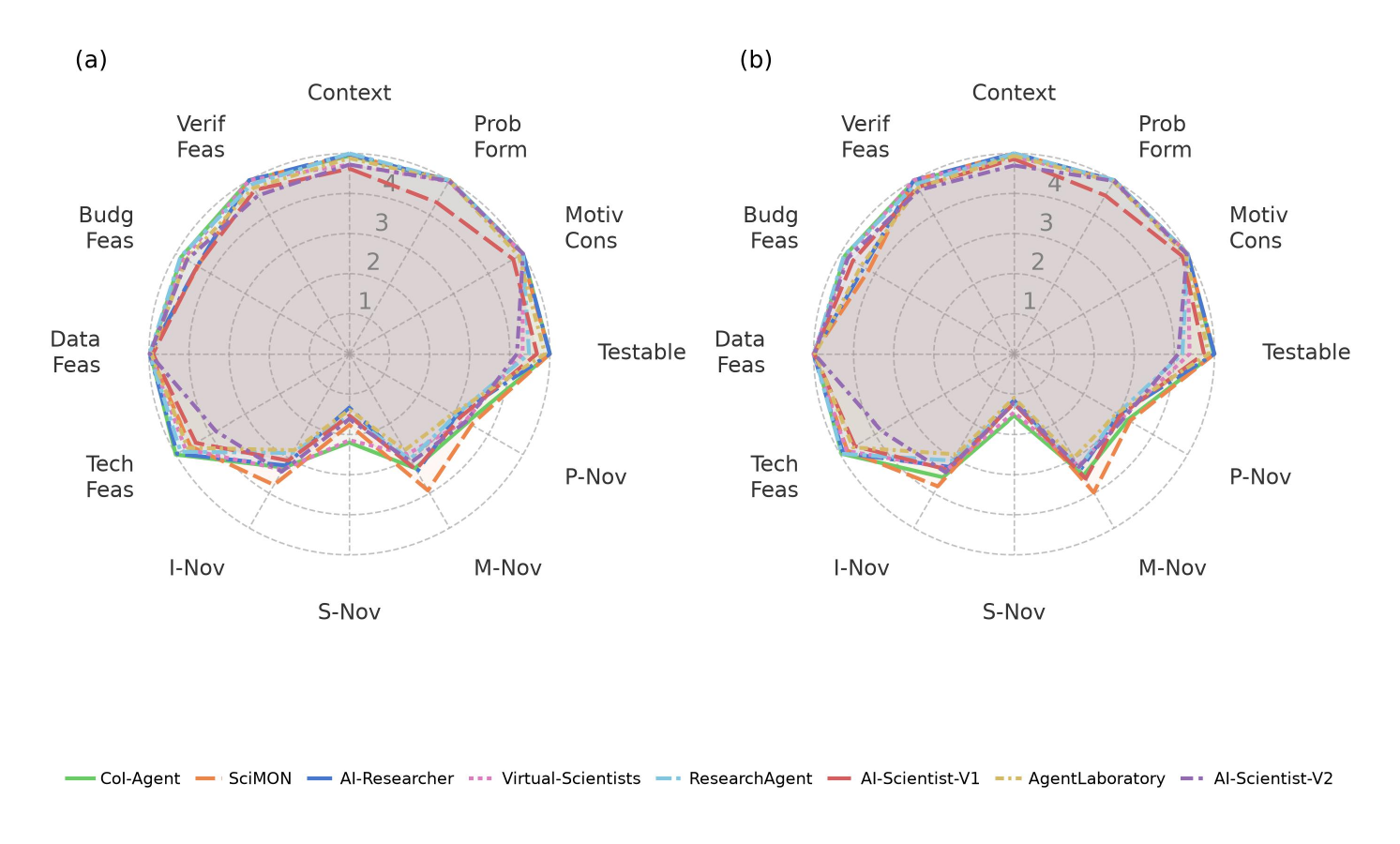}
\end{center}
\caption{Workshop versus Survey means pooled across agents. The two tracks almost coincide except on the novelty axes, where Workshop sits slightly farther from the center.}
\label{fig:modality_radar}
\end{figure}

\begin{figure*}[t]
\begin{center}
\includegraphics[width=\textwidth]{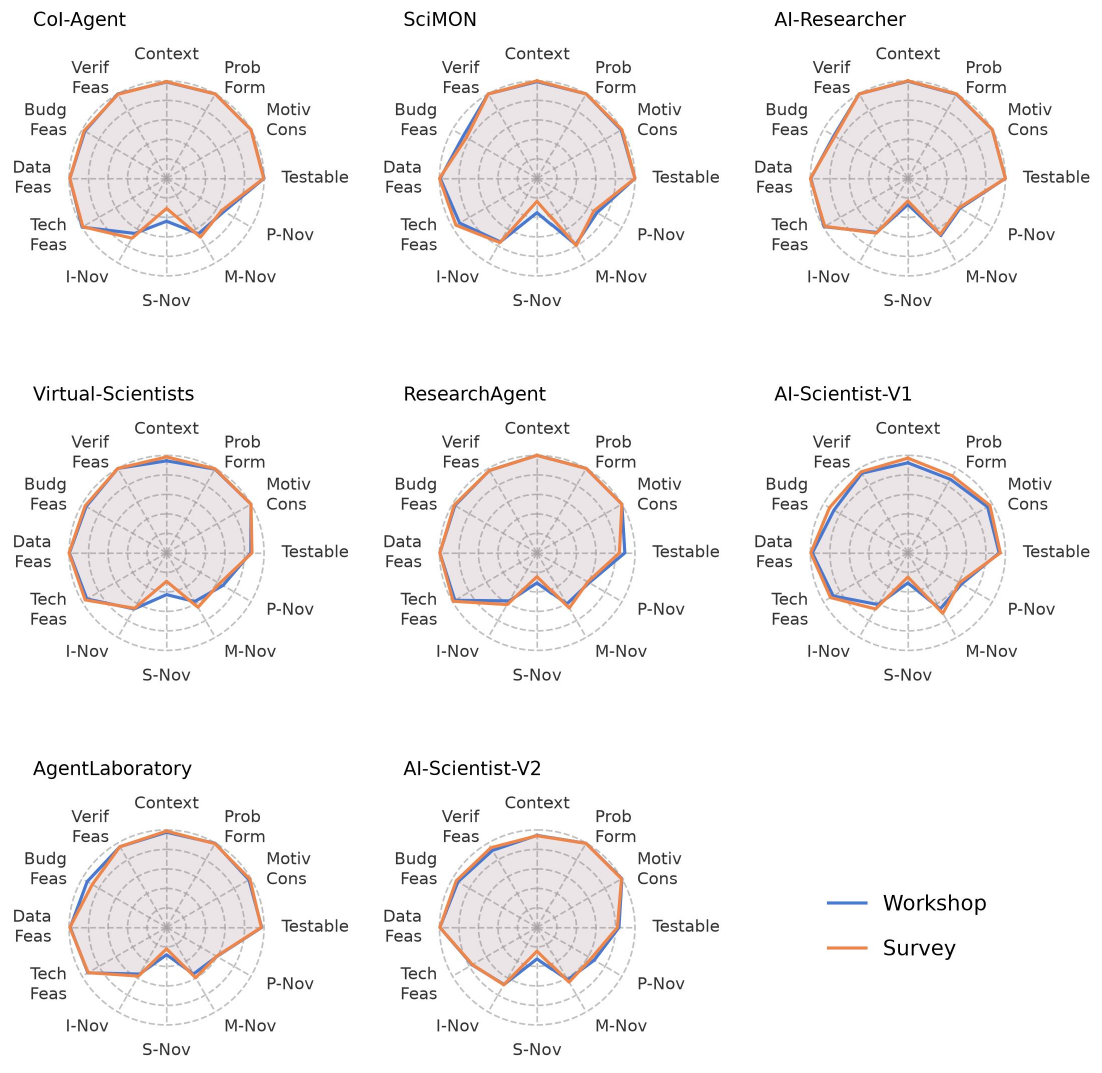}
\end{center}
\caption{Per agent Workshop versus Survey radars.}
\label{fig:modalities_grid}
\end{figure*}

Figure~\ref{fig:modality_radar} overlays Workshop and Survey means pooled across agents. The two tracks almost coincide except on the novelty axes. Figure~\ref{fig:modalities_grid} repeats that overlay once per agent. 

\clearpage

\section{Case study: seven per topic radars}
\label{app:casestudy}

\input{casestudy_appendix.tex}

\section{Budget envelope ablation on survey\_19}
\label{app:budget_ablation}

\begin{figure}[h]
\begin{center}
\includegraphics[width=0.92\linewidth]{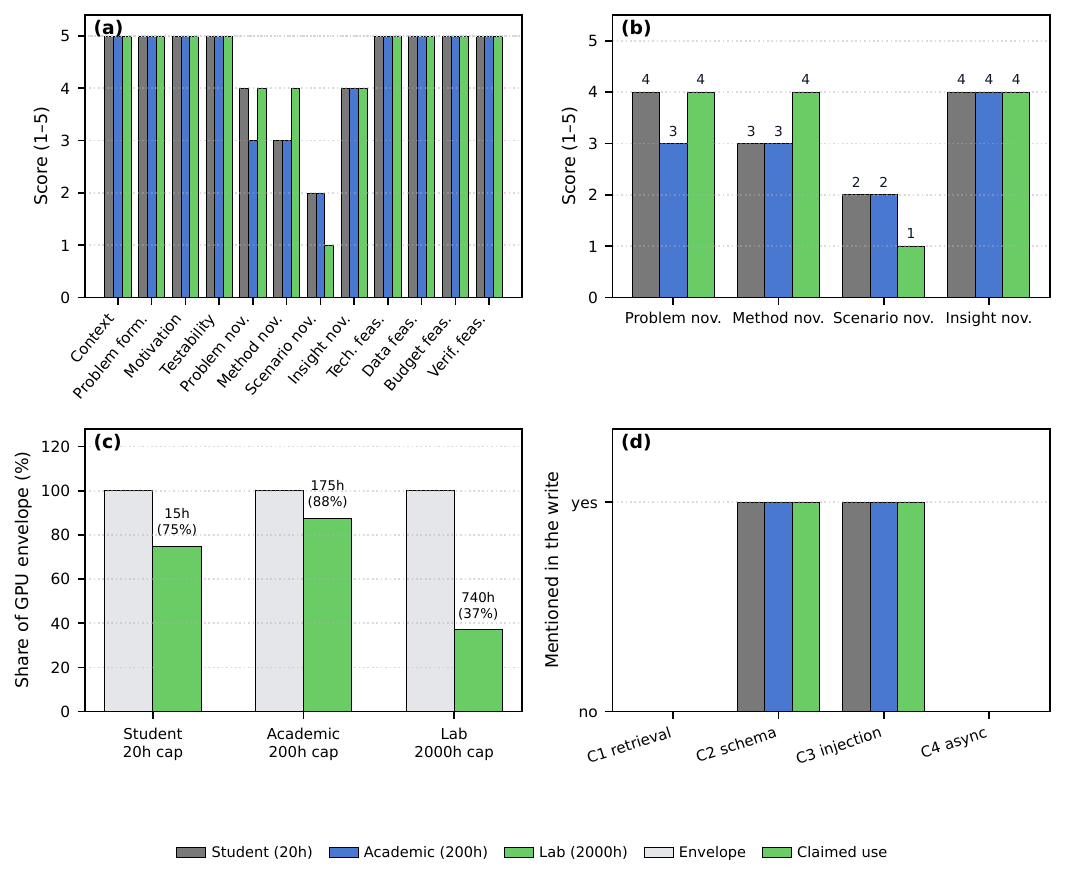}
\end{center}
\caption{CoI-Agent on survey\_19 under three GPU envelopes, same Gemini 3.7 Flash judge.
(a)~12 dimensions. (b)~Novelty axes, the only scores that move.
(c)~Claimed GPU hours versus the cap (log scale).
(d)~Which listed challenges the write names. Academic is TaintTool-Graph, not AsyncTx-Agent.}
\label{fig:budget_ablation_survey19}
\end{figure}

Figure~\ref{fig:budget_ablation} reports three topics. Figure~\ref{fig:budget_ablation_survey19} opens only the cheap tool-use survey. Validity and feasibility sit at 5 in every envelope, so ranking is a novelty ranking. After the same-200\,h rerun the academic write (TaintTool-Graph) stays on C2 and C3 and no longer touches C4; claimed compute is 175 of 200 GPU hours. The original AsyncTx write in Figure~\ref{fig:casestudy_box} is kept in the eight-family case study and is not replotted here.

\section{Ideation token use on survey\_19}
\label{app:ideation_tokens}

Table~\ref{tab:ideation_tokens} is one Gemini 3.7 Flash ideation run on \emph{survey\_19}, not a 100-task mean. Calls is the number of chat completions. Prompt and completion are the API usage fields; Total can exceed their sum because Gemini also reports thinking tokens.

\begin{table}[ht]
\caption{Chat token use for one Gemini 3.7 Flash ideation run on \emph{survey\_19}. Not a 100-task mean. Calls is the number of chat completions.}
\label{tab:ideation_tokens}
\begin{center}
\small
\begin{tabular}{@{}lrrrr@{}}
\toprule
Harness & Calls & Prompt & Completion & Total \\
\midrule
CoI-Agent & 23 & 104{,}483 & 24{,}748 & 141{,}792 \\
SciMON & 5 & 25{,}716 & 10{,}907 & 38{,}886 \\
AI-Researcher & 4 & 20{,}117 & 11{,}220 & 33{,}218 \\
Virtual-Scientists & 17 & 64{,}476 & 16{,}969 & 91{,}074 \\
ResearchAgent & 50 & 220{,}019 & 33{,}236 & 279{,}759 \\
AI-Scientist-V1 & 4 & 11{,}407 & 3{,}618 & 18{,}413 \\
AgentLaboratory & 5 & 27{,}163 & 12{,}114 & 42{,}188 \\
AI-Scientist-V2 & 2 & 6{,}672 & 1{,}497 & 9{,}968 \\
\bottomrule
\end{tabular}
\end{center}
\end{table}

\section{Judge protocol notes}
\label{app:judge}
The production novelty judge used in meta evaluation converts each paper to a structured idea object from title and abstract and never receives the official reviews. Agent scoring uses the 1 to 5 descriptors and checklists in Appendix~\ref{app:dimensions} on the generated JSON fields. Repeats per dimension in the 100 task study are 1.  Table~\ref{tab:coi_backbone} swaps only the CoI ideation model and keeps the Gemini 3.7 Flash judge.

\section{Ranking stability under a second judge}
\label{app:judge_agreement}

Table~\ref{tab:judge_agreement} compares Gemini 3.7 Flash with GPT-5 mini on the frozen ideas. Winner is the ideation system, not the judge. On the backbone row the two names are the Gemini-judge winner then the GPT-5 mini-judge winner.

\begin{table}[ht]
\caption{Ranking stability. Rank similarity when Gemini 3.7 Flash and GPT-5 mini are the judges. Top-3 overlap is $|A\cap B|/3$. Spearman $\rho$ is over the eight system ranks.}
\label{tab:judge_agreement}
\begin{center}
\scriptsize
\setlength{\tabcolsep}{4pt}
\begin{tabular}{@{}lcccc@{}}
\toprule
Evaluation target & Top-1 & Top-3 & Spearman $\rho$ & Winner \\
\midrule
Harness (Table~\ref{tab:macro_performance}) & Agree & 66.7\% (2/3) & 0.81 ($p{=}.015$) & CoI-Agent \\
CoI backbone (Table~\ref{tab:coi_backbone}) & Disagree & 66.7\% (2/3) & 0.83 ($p{=}.010$) & Gemini 3.7 Flash / GPT-5.4 mini \\
\bottomrule
\end{tabular}
\end{center}
\end{table}

\end{document}

%% file: math_commands.tex
\usepackage{amsmath,amsfonts,bm}

\def\eqref#1{equation~\ref{#1}}
\def\1{\bm{1}}

\DeclareMathAlphabet{\mathsfit}{\encodingdefault}{\sfdefault}{m}{sl}
\SetMathAlphabet{\mathsfit}{bold}{\encodingdefault}{\sfdefault}{bx}{n}

%% file: scoring_rubrics.tex
The Gemini 3.7 Flash judge scores one dimension at a time on an integer scale from 1 to 5 and must answer every checklist item as yes, no, partial, or n/a. A yes on a positive item supports a higher score; a yes on a negative item supports a lower score. Resource shortfalls are scored only under budget feasibility; they are not re-penalized in verification feasibility or research readiness. On novelty axes, a solid conference paper baseline is typically 3; 4 requires a clearly non-obvious new mechanism or formulation; 2 is routine module combination or minor adaptation. The descriptors below are the production rubrics.

\newcommand{\scoretab}[1]{%
\begin{center}
\small
\setlength{\tabcolsep}{4pt}
\begin{tabular}{@{}cp{0.88\linewidth}@{}}
\toprule
\# & Descriptor \\
\midrule
#1
\bottomrule
\end{tabular}
\end{center}
}

\subsection{Validity}

\paragraph{Context grounding.}
Does the idea correctly understand and use the given research background, prior papers, and research direction?
\scoretab{
1 & Largely detached from the given context, or contains serious factual errors; the problem is unrelated to the given direction \\
2 & Superficially related to the context, but clearly misreads prior work, research goals, or limitations \\
3 & Broadly understands the context, but only uses the general theme and does not use the key papers or research gaps accurately \\
4 & Correctly understands prior work and identifies a research gap consistent with the context \\
5 & Deeply and accurately synthesises the papers, limitations, and background in the context; the idea follows naturally from that context \\
}
\textit{Positive checks.} Is the idea relevant to the given research direction. Are existing methods described accurately. Are the limitations of prior work identified correctly. Is key information from the context actually used.
\textit{Negative checks.} Is a problem already solved by prior work mistaken for an open problem. Are there factual errors about the papers. Does it introduce a key assumption absent from the context without explaining where it comes from. Are there fabricated papers, fabricated results, or incorrect citations.

\paragraph{Problem formulation.}
Does the idea turn a broad direction into a clear, specific, well-bounded research problem?
\scoretab{
1 & No clear research problem; only a topic, a slogan, or a wish \\
2 & The general direction is visible, but the subject, conditions, goal, and boundaries are all vague \\
3 & The problem is broadly understandable, but important qualifying conditions or key variables are still missing \\
4 & Problem, subject, applicable conditions, and goal are reasonably clear, and the scope is roughly manageable \\
5 & The research problem is defined precisely, with clear key variables, boundaries, assumptions, and goals; it could almost be turned directly into a formal research question \\
}
\textit{Positive checks.} Is the object of study stated. Is the phenomenon or failure under study stated. Are the study conditions stated. Are the key variables stated. Is the research goal stated. Is there an explicit research question or hypothesis.
\textit{Negative checks.} Is it merely vague phrasing such as ``improve performance'', ``enhance capability'', or ``solve robustness''. Are several unrelated problems mixed into one idea. Is the scope of the idea too large.

\paragraph{Motivation and method consistency.}
Does the proposed method actually address the research gap identified in the motivation?
\scoretab{
1 & Method and motivation are essentially unrelated; there is no visible reason the method would solve the problem \\
2 & Method relates to the motivation only superficially and cannot address the core research gap \\
3 & Method broadly responds to the motivation, but a key mechanism or logical link in between is missing \\
4 & Method explicitly targets the core problem in the motivation, with a plausible mechanism of action \\
5 & Motivation, hypothesis, method, and experiments form a closed loop; the method addresses the core bottleneck rather than a peripheral issue \\
}
\textit{Positive checks.} Is the bottleneck identified by the motivation stated explicitly. Does the method target that bottleneck directly. Can the method test the key hypothesis. If the method succeeds, would the original problem be alleviated. Would the planned experiments show that the method really solves the problem in the motivation.
\textit{Negative checks.} Is there a ``motivation says A, method does B'' mismatch. Is the method merely a module related to the problem without actually solving it.

\paragraph{Testability.}
Can the core scientific claim of this idea be clearly tested or falsified?
\scoretab{
1 & No clear way to verify it; the core claim is essentially untestable \\
2 & There is a rough experimental direction, but no clear variables, metrics, or decision criteria \\
3 & A preliminary test is possible, but strong baselines, controls, or clear falsification criteria are missing \\
4 & Clear hypothesis, experimental setup, evaluation metrics, and comparison targets \\
5 & Rigorous falsifiable hypothesis, controlled variables, control conditions, success criteria, and analysis of alternative explanations \\
}
\textit{Positive checks.} Is there an explicit hypothesis. Are there observable variables. Is there a measurable outcome. Is there a baseline. Is there a control condition. Is there an explicit success criterion. Could the results support or overturn the core claim. Can competing explanations be distinguished. Is ``the model gets better'' turned into a measurable metric.
\textit{Negative checks.} Are there untestable strong claims such as ``truly understands'' or ``achieves general intelligence''.

\subsection{Novelty}

\paragraph{Problem novelty.}
Is the research problem, phenomenon, failure mode, hypothesis, or research perspective new relative to prior work?
\scoretab{
1 & The problem has already been explicitly posed and largely solved in prior literature \\
2 & Minor repackaging, cosmetic re-framing, or routine extension of an existing problem without posing a new fundamental question \\
3 & Meaningful new angle or specific failure mode within an existing problem domain, but closely following established research trajectories \\
4 & Distinctly new, well-defined research problem, non-obvious failure mode, or mechanistic question that challenges existing assumptions \\
5 & Foundational new problem formulation or conceptual paradigm that defines an entirely new research area \\
}
\textit{Positive checks.} Does it identify a genuinely new failure mode or unexplored bottleneck. Does it identify a new contradiction or challenge to existing assumptions. Does it pose a distinct, non-obvious research question. Does it propose a new problem boundary or object of study. Is there a substantive difference from closest prior work.
\textit{Negative checks.} Has this problem already been studied or largely solved in existing literature. Is it only a cosmetic re-framing or routine change of dataset, model, or benchmark.

\paragraph{Method novelty.}
Is the algorithm, model, training procedure, experimental method, or theoretical tool used to solve the problem new?
\scoretab{
1 & The method essentially restates, directly adopts, or replicates existing methods with no substantive modification \\
2 & Routine combination or minor adaptation: standard stitching of off-the-shelf modules, trivial backbone swap, or straightforward application of existing toolkits \\
3 & Solid incremental method: meaningful procedural or architectural modification tailored to the problem, but the core mechanism remains within established paradigms \\
4 & Non-obvious, substantive new method design: introduces a distinct, well-formulated new algorithmic formulation, loss objective, mechanism, or theoretical tool \\
5 & Foundational or paradigm-shifting methodological breakthrough that establishes a fundamentally new computing, modeling, or learning principle \\
}
\textit{Positive checks.} Does the method introduce a non-obvious algorithmic delta beyond standard module combinations. Does it change the core solving mechanism or mathematical or algorithmic formulation. Is the new method specifically and non-trivially designed for the bottleneck of this idea.
\textit{Negative checks.} Does the method already appear in existing papers or standard toolkits. Is it primarily a routine assembly or stitching of standard existing components (for example standard models, prompting, or off-the-shelf RL or loss). Is it only a backbone swap, hyperparameter tuning, or dataset swap. Is it only the addition of a common module without changing the core solving mechanism. If it is a combination of methods, does that combination already have direct precedent or is it a routine pairing (answer n/a if the method is not a combination).

\paragraph{Scenario novelty.}
Does the idea introduce a new domain, task, modality, data distribution, deployment environment, or real-world constraint, and does that give rise to a meaningful new research problem?
\scoretab{
1 & The scenario completely follows existing work, or is only renamed \\
2 & Only the dataset, application domain, or input format changes; the core research problem and technical challenges are essentially unchanged \\
3 & Introduces a new application domain, task, or constraint, but standard methods can be transferred directly without fundamental adaptations \\
4 & The new scenario creates distinct, meaningful challenges that fundamentally require targeted method redesign rather than direct transfer \\
5 & The new scenario opens an entirely new research frontier with unprecedented evaluation, modeling, or real-world requirements \\
}
\textit{Positive checks.} Is it a new domain, task, modality, or deployment environment. Does the new scenario introduce unique challenges that prevent standard direct transfer. Does the scenario necessitate targeted algorithmic or evaluation redesign. Does the new scenario change the underlying research problem.
\textit{Negative checks.} Is it just applying existing methods directly to a new dataset or domain without unique challenges.

\paragraph{Insight novelty.}
Does the idea offer a new understanding of a phenomenon, failure mode, method behaviour, or causal mechanism?
\scoretab{
1 & No new explanation; only describes a phenomenon, repeats known conclusions, or provides cosmetic explanations \\
2 & Offers a generic, high-level, or templated explanation that does not reveal concrete mechanisms \\
3 & Provides a specific empirical observation or preliminary mechanistic hypothesis, but remains close to existing understandings \\
4 & Proposes a clear, non-obvious new mechanism, boundary condition, or causal explanation supported by concrete evidence or falsifiable claims \\
5 & Reveals a foundational, non-obvious new principle or causal mechanism that alters how the field understands the phenomenon \\
}
\textit{Positive checks.} Does it propose a new mechanistic or causal explanation. Does it point to specific, verifiable causal factors. Does it identify a new failure mechanism or boundary condition. Does it disentangle factors previously conflated. Is it substantively different from existing explanations in the literature. Can the mechanistic claim be tested experimentally or theoretically.
\textit{Negative checks.} Is it merely renaming an existing phenomenon or offering a generic explanation.

\subsection{Feasibility}

\paragraph{Technical feasibility.}
Can the technical plan of this idea be implemented?
\scoretab{
1 & Contains a fundamental technical contradiction, or depends on unobtainable information; essentially not implementable \\
2 & Key technical components are missing; only a conceptual description, hard to implement \\
3 & A simplified version or prototype is implementable, but the core plan still carries clear technical risk \\
4 & The technical path is clear, the required components mostly exist, and it can reasonably be implemented \\
5 & The implementation path is mature and complete, technical risk is low, and prototyping could start immediately \\
}
\textit{Positive checks.} Do the required algorithms, models, and components exist. Is the method pipeline complete. Are inputs and outputs clearly defined. Are the training and inference loops closed. Can a prototype be built.
\textit{Negative checks.} Does it depend on unobtainable information. Does it require an oracle. Are there self-contradictory assumptions. Does it require solving an unmentioned core technical challenge.

\paragraph{Data feasibility.}
Can the data, labels, ground truth, and data access this idea needs be obtained?
\scoretab{
1 & The required data does not exist, cannot be obtained, or there is no workable ground truth \\
2 & Major obstacles to obtaining data; requires large-scale collection, expert annotation, or access approval \\
3 & Substitute data or a simplified version is possible, but data quality, scale, or representativeness is limited \\
4 & The required data is largely obtainable, needing only routine cleaning, processing, or limited annotation \\
5 & Data is public, sufficient, of suitable quality, with reliable labels, and can directly support verification \\
}
\textit{Positive checks.} Does the data exist. Is it publicly available. Is the amount of data sufficient. Is the data quality sufficient. Is there reliable ground truth. Can train, validation, and test splits be constructed. Can the data really support the claim being verified.
\textit{Negative checks.} Does new data need to be collected. Is expert annotation required. Are there privacy, copyright, or ethical issues. Could train-test leakage occur.

\paragraph{Budget feasibility.}
Can this idea be completed under the compute, time, API, human-effort, and funding constraints specified by the benchmark?
\scoretab{
1 & Clearly exceeds the stated budget; no meaningful verification is possible as things stand \\
2 & Key resources are severely insufficient; it can only remain conceptual \\
3 & A scaled-down proof-of-concept is possible, but full verification is not \\
4 & Reasonable experiments can be completed within the given compute, time, and funding \\
5 & Resource needs are clearly below budget, the experimental path is clear, and there is ample headroom for baselines, ablations, and repeated runs \\
}
\textit{Positive checks.} Is GPU or CPU capacity sufficient. Is training time sufficient. Is inference cost affordable. Is the human annotation cost affordable. Is there enough time for data collection and processing. Is there enough time for baselines, ablations, and repeated runs. Can a meaningful verification be completed within budget, not just a working demo.
\textit{Negative checks.} Does the number of API calls exceed the budget. Are special devices or a deployment environment required.

\paragraph{Verification feasibility.}
Under real conditions, can a sufficiently credible experimental or theoretical verification be carried out?
\scoretab{
1 & Credible verification is essentially impossible; data, metrics, baselines, or observable outcomes are missing \\
2 & Only indirect, insufficient, or highly unreliable evidence can be obtained \\
3 & Preliminary verification is possible, but experiment scale, controls, or measurement capability are limited \\
4 & Clear, executable experimental procedure, baselines, metrics, and basic controls \\
5 & A rigorous, reproducible verification plan that can distinguish alternative explanations and support falsification \\
}
\textit{Positive checks.} Is there an executable experimental procedure. Is the required data obtainable. Are baselines implemented. Are the evaluation metrics appropriate. Can ablations be carried out. Can key variables be controlled. Can experiments be repeated. Can alternative explanations be ruled out. Is the resulting evidence strong enough to be credible rather than merely suggestive. Could the results support or refute the core claim. Whether resources fit the budget belongs to budget feasibility; do not re-penalize the budget here.

%% file: casestudy_appendix.tex
This appendix is the reviewer facing case study for Figure~\ref{fig:casestudy_radar}. For each of the seven topics we give the packaged input, a 12 dimension score table, a one paragraph diagnosis of why the scores move, and a length capped excerpt of each family's idea. Question fields are cut at 180 characters and method fields at 280; `[...]` marks a cut. Experiments, budgets, and the 12 dimension judge comments are in the supplementary PDF. Full JSON and checklists are in the supplement.

The reading rule is the same on every panel. Validity hugs the outer ring because the agents fill the JSON template. Overall means near 4.3 can hide a one idea cluster. Scenario novelty is what marks a family that leaves the cluster. Which family that is depends on the input frame, not on a stable global ranking.

\subsection{The seven topics together}
On the agentic CFP, {SciMON} is the scenario outlier and the others collapse onto a Popperian science agent. On the tool learning survey, {CoI-Agent} is the scenario outlier and the others collapse onto prompt injection. On drug modalities, four families leave the protein pocket default. On generative biology, only Virtual Scientists change the operating constraint, and they pay for it in budget. On geometry, incomplete JSON is enough to collapse {AI-Scientist-V1} and {AgentLaboratory}. We do not claim these topics are typical of all 100 tasks.

\subsection{Workshop: agentic AI for science}

\emph{ws\_02\_agentic\_ai.}

The input is the ICLR 2025 CFP \emph{Towards Agentic AI for Science: Hypothesis Generation, Comprehension, Quantification, and Validation}. It lists four open thrusts (system design, theory, applications, and validation) and does not name a single fashionable bottleneck.

\begin{table}[h]
\caption{All 12 dimension scores on \emph{ws\_02\_agentic\_ai}. Highest in each row is bold.}
\label{tab:casestudy_ws_scores}
\begin{center}
\small
\setlength{\tabcolsep}{3.6pt}
\resizebox{\linewidth}{!}{%
\begin{tabular}{lcccccccc}
\toprule
Dimension & \shortstack{CoI-\\Agent} & SciMON & \shortstack{AI-\\Researcher} & \shortstack{Virtual-\\Scientists} & \shortstack{Research\\Agent} & \shortstack{AI-Scientist-\\V1} & \shortstack{Agent\\Laboratory} & \shortstack{AI-Scientist-\\V2} \\
\midrule
Context grounding & \textbf{5} & \textbf{5} & \textbf{5} & \textbf{5} & \textbf{5} & \textbf{5} & \textbf{5} & \textbf{5} \\
Problem formulation & \textbf{5} & \textbf{5} & \textbf{5} & 4 & \textbf{5} & \textbf{5} & \textbf{5} & \textbf{5} \\
Motivation consistency & \textbf{5} & \textbf{5} & \textbf{5} & \textbf{5} & \textbf{5} & \textbf{5} & \textbf{5} & \textbf{5} \\
Testability & \textbf{5} & \textbf{5} & \textbf{5} & 4 & \textbf{5} & \textbf{5} & \textbf{5} & 4 \\
Problem novelty & \textbf{4} & \textbf{4} & \textbf{4} & 3 & 3 & \textbf{4} & 3 & 3 \\
Method novelty & 3 & \textbf{4} & \textbf{4} & 2 & 3 & 3 & 3 & 3 \\
Scenario novelty & 3 & \textbf{4} & 2 & 3 & 2 & 2 & 1 & 2 \\
Insight novelty & 3 & \textbf{4} & \textbf{4} & 2 & 3 & 3 & 3 & 3 \\
Technical feasibility & \textbf{5} & \textbf{5} & \textbf{5} & \textbf{5} & \textbf{5} & \textbf{5} & 4 & 3 \\
Data feasibility & \textbf{5} & \textbf{5} & \textbf{5} & \textbf{5} & \textbf{5} & \textbf{5} & \textbf{5} & \textbf{5} \\
Budget feasibility & \textbf{5} & \textbf{5} & \textbf{5} & \textbf{5} & \textbf{5} & \textbf{5} & \textbf{5} & 4 \\
Verification feasibility & \textbf{5} & \textbf{5} & \textbf{5} & \textbf{5} & \textbf{5} & \textbf{5} & \textbf{5} & \textbf{5} \\
\bottomrule\end{tabular}%
}
\end{center}\end{table}
\paragraph{Why the scores move.}
Six families write a variant of ``make the science agent falsify its own hypothesis''. The labels differ (severe testing, minimax skeptics, conformal checks, information gain), but the operating setting is still an LLM science agent with a debate or verifier loop, so scenario novelty stays at 1 to 3. {SciMON} is the only system at 4 on all four novelty axes: CausalPopper forces an identifiable structural causal model and a contrastive intervention, which the judge treats as a change of scientific environment rather than another debate template. {CoI-Agent} also reaches chemistry tools and optimal design (scenario 3) but keeps method novelty at 3 because the core loop is still competing DAGs plus Bayesian design. Virtual Scientists leave the generic science agent and audit medical imaging workflows (scenario 3), at the cost of method and insight (both 2): the judge reads ARCA as a calibration and confounder-stress toolkit stitched onto existing radiology agents. {AgentLaboratory} stays inside the Popper cluster on LAB-Bench and ChemBench (scenario 1). {AI-Scientist-V2} is the brittle case: the judge says the core workflow can be prototyped, but the write is underspecified, so technical feasibility drops to 3 and testability to 4. Validity remains at ceiling for every family except those two execution dents.

\paragraph{Idea excerpts.} Question at most 180 characters, method at most 280; longer fields are cut. Judge comments are in the supplementary PDF.

\begin{sloppypar}\small
\noindent\textbf{CoI-Agent: AutoCausal-Sci.} \textit{M.} AutoCausal-Sci, a closed-loop multi-agent architecture that frames scientific hypothesis exploration as Active Causal Disambiguation with Bayesian Optimal Experimental Design (OED) under rigorous sequential error bounds. [...]

\smallskip
\noindent\textbf{SciMON: CausalPopper.} \textit{Q.} How can autonomous agentic AI systems transition from generating narrative, plausibility-driven scientific ideas to producing provably identifiable Structural Causal Models (SCMs) [...] \textit{M.} CausalPopper is an end-to-end framework consisting of four core components: 1. Non-Parametric SCM Hypothesis Representation: Formalizes candidate scientific hypotheses as M k = V, U k, F k, P(U k) , where V are observed scientific variables, U k are explicit latent confounders, [...]

\smallskip
\noindent\textbf{AI-Researcher: PAD.} \textit{Q.} How can autonomous multi-agent Large Language Model (LLM) architectures be formulated to generate, quantify, and validate scientific hypotheses without succumbing to degenerate con [...] \textit{M.} Popperian Agentic Discovery (PAD) formalizes scientific discovery as an asymmetric minimax epistemic game coupled with neuro-symbolic reasoning and multi-fidelity Bayesian active learning: 1. [...]

\smallskip
\noindent\textbf{Virtual-Scientists: ARCA.} \textit{M.} ARCA (Adversarial Reliability and Calibration Auditor), a principled, modular auditing and verification framework designed to stress-test, calibrate, and falsify agentic scientific discovery in medical imaging. [...]

\smallskip
\noindent\textbf{ResearchAgent: CO-DISCOVER.} \textit{Q.} How can we design a resource-bounded, multi-agent scientific discovery framework - integrating specialized hypothesis generation, physical/domain simulation tools, and epistemic unce [...] \textit{M.} CO-DISCOVER: Cost-Optimal Epistemic Discovery via Multi-Agent Active Inference and Multi-Fidelity Validation CO-DISCOVER is a resource-bounded, multi-agent scientific discovery framework that formulates the cycle of hypothesis generation, quantification, validation, and refinemen [...]

\smallskip
\noindent\textbf{AI-Scientist-V1: Active Dialectical Falsification.} \textit{Q.} Can an asymmetric minimax game between an inductive hypothesis generator and an adversarial falsification agent significantly improve hypothesis precision and sample efficiency in [...] \textit{M.} The Active Dialectical Falsification (ADF) framework operates as a discrete-round asymmetric minimax game composed of four specialized agent modules: 1. [...]

\smallskip
\noindent\textbf{AgentLaboratory: CNS-Agent.} \textit{Q.} How can autonomous multi-agent scientific discovery systems eliminate compounding hallucinations and ungrounded consensus in multi-step hypothesis generation by integrating abducti [...] \textit{M.} The CNS-Agent framework integrates three tightly coupled modules: (1) Abductive Proposer mathcalM textprop (using LLaMA-3.1-8B-Instruct and Mistral-NeMo-12B) which decomposes scientific research queries into atomic sub-hypotheses and candidate protocol trees; (2) Determ [...]

\smallskip
\noindent\textbf{AI-Scientist-V2: AutoPopper.} \textit{Q.} Autonomous scientific AI agents suffer from pervasive confirmation bias, generating safe or unfalsifiable hypotheses and executing experimental designs that fail to distinguish the [...] \textit{M.} AutoPopper, a multi-agent framework grounded in the philosophy of Severe Testing and epistemic calibration for autonomous scientific discovery. [...]
\end{sloppypar}

\subsection{Survey: tool learning}

\emph{survey\_19\_agentic\_tool\_use.}

The input is the survey \emph{Tool Learning with Large Language Models} and four author stated bottlenecks: (C1) retrieval among thousands of APIs, (C2) nested schema and parameter compliance, (C3) indirect prompt injection from tool outputs, and (C4) concurrent asynchronous orchestration.

\begin{table}[h]
\caption{All 12 dimension scores on \emph{survey\_19\_agentic\_tool\_use}. Highest in each row is bold.}
\label{tab:casestudy_scores}
\begin{center}
\small
\setlength{\tabcolsep}{3.6pt}
\resizebox{\linewidth}{!}{%
\begin{tabular}{lcccccccc}
\toprule
Dimension & \shortstack{CoI-\\Agent} & SciMON & \shortstack{AI-\\Researcher} & \shortstack{Virtual-\\Scientists} & \shortstack{Research\\Agent} & \shortstack{AI-Scientist-\\V1} & \shortstack{Agent\\Laboratory} & \shortstack{AI-Scientist-\\V2} \\
\midrule
Context grounding & \textbf{5} & \textbf{5} & \textbf{5} & \textbf{5} & \textbf{5} & \textbf{5} & \textbf{5} & \textbf{5} \\
Problem formulation & \textbf{5} & \textbf{5} & \textbf{5} & \textbf{5} & \textbf{5} & \textbf{5} & \textbf{5} & \textbf{5} \\
Motivation consistency & \textbf{5} & \textbf{5} & \textbf{5} & \textbf{5} & \textbf{5} & \textbf{5} & \textbf{5} & \textbf{5} \\
Testability & \textbf{5} & \textbf{5} & \textbf{5} & 4 & \textbf{5} & \textbf{5} & \textbf{5} & \textbf{5} \\
Problem novelty & \textbf{4} & \textbf{4} & 3 & \textbf{4} & 3 & 3 & 3 & 3 \\
Method novelty & \textbf{4} & \textbf{4} & \textbf{4} & 3 & \textbf{4} & 3 & \textbf{4} & 3 \\
Scenario novelty & \textbf{4} & 2 & 1 & 2 & 1 & 2 & 1 & 1 \\
Insight novelty & \textbf{4} & \textbf{4} & \textbf{4} & \textbf{4} & \textbf{4} & 3 & \textbf{4} & 3 \\
Technical feasibility & \textbf{5} & \textbf{5} & \textbf{5} & \textbf{5} & \textbf{5} & 4 & \textbf{5} & \textbf{5} \\
Data feasibility & 4 & \textbf{5} & \textbf{5} & \textbf{5} & \textbf{5} & \textbf{5} & \textbf{5} & \textbf{5} \\
Budget feasibility & \textbf{5} & 4 & \textbf{5} & \textbf{5} & \textbf{5} & \textbf{5} & \textbf{5} & \textbf{5} \\
Verification feasibility & \textbf{5} & \textbf{5} & \textbf{5} & \textbf{5} & \textbf{5} & \textbf{5} & \textbf{5} & \textbf{5} \\
\bottomrule\end{tabular}%
}
\end{center}\end{table}
\paragraph{Why the scores move.}
Five families independently lock onto C3. Their titles all advertise a control versus data split: asymmetric attention, dual stream privilege, taint tags, or a plan frozen before untrusted observations enter the context. The judge marks method novelty 3 or 4 because the gadget differs, but scenario novelty stays at 1 or 2 because the evaluation plans reuse InjecAgent, BIPIA, BFCL, and ToolBench. {SciMON} is the only system that ignores injection and attacks C2: SynApse compiles OpenAPI schemas into a semantic attributed pushdown automaton. Method and insight are 4; scenario is 2 because the setting is still REST or OpenAPI function calling. {CoI-Agent} is the only proposal that treats C4 as first class. AsyncTx-Agent models concurrent calls as transactions with speculative rollback; the judge gives scenario 4 and still keeps testability and budget at 5 (7B or 8B LoRA, itemized GPU hours). A same-200\,h CoI rerun on this survey writes TaintTool-Graph instead (scenario 2; Section~\ref{subsec:budget_ablation}), so the scenario-4 cell here is not a stable envelope effect. Virtual Scientists sit between the clusters (CFG decoding plus a sandbox), so method novelty drops to 3 and testability to 4. On this topic the ranking is not ``everyone is around 4.3'': {CoI-Agent} changes the operating constraint, {SciMON} changes the decoder, and the other six are close variants of one security story.

\paragraph{Idea excerpts.} Question at most 180 characters, method at most 280; longer fields are cut. Judge comments are in the supplementary PDF.

\begin{sloppypar}\small
\noindent\textbf{CoI-Agent: AsyncTx-Agent.} \textit{(method field empty in the released JSON)}

\smallskip
\noindent\textbf{SciMON: SynApse.} \textit{Q.} How can large language models guarantee 100\% syntactic and cross-field relational invariant compliance in deep polymorphic tool schemas without rejection sampling, while optimizing [...] \textit{M.} SynApse consists of three tightly coupled innovations: 1. Semantic Attributed Pushdown Automaton (S-APDA): Defined as M = (Q, Sigma, Gamma, delta, q 0, Z 0, A) with attribute tuple A = (I, S, F). [...]

\smallskip
\noindent\textbf{AI-Researcher: SECURE-TOOL.} \textit{Q.} How can open-weight foundation models achieve provable structural resistance against indirect prompt injection embedded in external tool observations while concurrently eliminating [...] \textit{M.} SECURE-TOOL introduces a dual-plane security and alignment architecture consisting of: (1) Token Plane Partitioning separating token indices into Control tokens C (system directives, user instructions, thoughts), Untrusted Tool Observations D, and Action Tokens A. [...]

\smallskip
\noindent\textbf{Virtual-Scientists: SafeSchema.} \textit{M.} SafeSchema resolves both challenges via an integrated, mathematically grounded dual-track execution framework: 1. Dual-Track Content-Addressed Sandboxing (Control/Data Decoupling): SafeSchema physically decouples the primary Agent Instruction \& Planning Stream from raw observatio [...]

\smallskip
\noindent\textbf{ResearchAgent: Privilege-Gated Asymmetric Cross-Attention.} \textit{Q.} How can large language model (LLM) agents maintain operational integrity against indirect prompt injection embedded within untrusted external tool responses while preserving task r [...] \textit{M.} Privilege-Gated Asymmetric Cross-Attention: The control stream attends to the data stream exclusively through an Asymmetric Privilege Adapter (APA) - a low-rank cross-attention layer integrated into transformer decoder blocks. [...]

\smallskip
\noindent\textbf{AI-Scientist-V1: SecurTool.} \textit{Q.} How can multi-turn LLM tool-use architectures prevent indirect prompt injection attacks embedded in untrusted tool observations from hijacking high-privilege execution arguments wi [...] \textit{M.} SecurTool introduces a dual-phase defense combining an architectural parameter separation mechanism and provenance-grounded alignment: 1. [...]

\smallskip
\noindent\textbf{AgentLaboratory: Taint-Aware Dual-Stream Attention.} \textit{Q.} Can architectural disentanglement of control-plane planning attention heads and data-plane extraction attention heads in open-weight Large Language Models eliminate Indirect Tool O [...] \textit{M.} Taint-Aware Dual-Stream Attention (TADSA) introduces explicit taint-tracking and dual-stream attention routing into open-weight transformer architectures (e.g., Llama-3-8B-Instruct, Mistral-7B-v0.3): 1. [...]

\smallskip
\noindent\textbf{AI-Scientist-V2: IntentLock.} \textit{Q.} Indirect prompt injection (IPI) vulnerabilities in tool-equipped LLMs stem from the unconstrained mingling of untrusted tool observations with the agent's autoregressive control-fl [...] \textit{M.} IntentLock, an architectural and inference-time orchestration framework that strictly separates Control-Flow Planning from Untrusted Data-Value Binding. [...]
\end{sloppypar}

\subsection{Workshop: AI for new drug modalities}

\emph{ws\_16\_ai\_for\_drug\_discovery.}

The input is the NeurIPS 2024 CFP \emph{AI for New Drug Modalities}. It asks for work that bridges AI with emerging modalities (RNA, gene, and cell therapies), protein engineering, and new molecular representations, rather than only classical small-molecule docking.

\begin{table}[h]
\caption{All 12 dimension scores on \emph{ws\_16\_ai\_for\_drug\_discovery}. Highest in each row is bold.}
\label{tab:cs_ws16aifordrugdiscovery}
\begin{center}
\small
\setlength{\tabcolsep}{3.6pt}
\resizebox{\linewidth}{!}{%
\begin{tabular}{lcccccccc}
\toprule
Dimension & \shortstack{CoI-\\Agent} & SciMON & \shortstack{AI-\\Researcher} & \shortstack{Virtual-\\Scientists} & \shortstack{Research\\Agent} & \shortstack{AI-Scientist-\\V1} & \shortstack{Agent\\Laboratory} & \shortstack{AI-Scientist-\\V2} \\
\midrule
Context grounding & \textbf{5} & \textbf{5} & \textbf{5} & 4 & \textbf{5} & \textbf{5} & 4 & \textbf{5} \\
Problem formulation & \textbf{5} & \textbf{5} & \textbf{5} & \textbf{5} & \textbf{5} & \textbf{5} & \textbf{5} & \textbf{5} \\
Motivation consistency & \textbf{5} & \textbf{5} & \textbf{5} & \textbf{5} & \textbf{5} & \textbf{5} & \textbf{5} & \textbf{5} \\
Testability & \textbf{5} & \textbf{5} & \textbf{5} & 4 & 4 & \textbf{5} & \textbf{5} & 4 \\
Problem novelty & 3 & 3 & 3 & \textbf{4} & 3 & 3 & 3 & \textbf{4} \\
Method novelty & 3 & \textbf{4} & 3 & 3 & 3 & 3 & 2 & 3 \\
Scenario novelty & \textbf{4} & 1 & 1 & \textbf{4} & \textbf{4} & 1 & 1 & \textbf{4} \\
Insight novelty & 3 & 3 & 2 & \textbf{4} & 2 & 3 & 2 & \textbf{4} \\
Technical feasibility & \textbf{5} & \textbf{5} & \textbf{5} & 4 & 4 & 4 & \textbf{5} & \textbf{5} \\
Data feasibility & \textbf{5} & \textbf{5} & \textbf{5} & \textbf{5} & \textbf{5} & \textbf{5} & \textbf{5} & \textbf{5} \\
Budget feasibility & 4 & \textbf{5} & \textbf{5} & \textbf{5} & \textbf{5} & \textbf{5} & 4 & \textbf{5} \\
Verification feasibility & \textbf{5} & \textbf{5} & \textbf{5} & \textbf{5} & 4 & \textbf{5} & \textbf{5} & \textbf{5} \\
\bottomrule\end{tabular}%
}
\end{center}\end{table}
\paragraph{Why the scores move.}
The CFP asks for new drug modalities. Four families actually leave the default protein-pocket SBDD setting and reach scenario 4: {AI-Scientist-V2} (molecular glues at bipartite E3 interfaces), Virtual Scientists (RNA-targeted ligands), {CoI-Agent} (modality-aware diffusion), and {ResearchAgent} (macrocyclic peptides). Those four are the ones that fill the novelty notch on this panel. {SciMON} again writes a stronger method (directional interaction tensor fields, method 4) inside CrossDocked and PDBbind (scenario 1). {AgentLaboratory} and {AI-Researcher} stay on protein-pocket PK-guided diffusion (scenario 1, novelty 2.00 to 2.25). Virtual Scientists pay a small testability and technical cost (both 4) for the RNA move; {AI-Scientist-V2} keeps technical feasibility at 5 on GlueDiff because the write stays inside existing SE(3) diffusion stacks. The lesson is the same split as on the tool survey, with the opposite winner: leaving the fashionable protein-pocket setup, not adding another guidance term, is what the scenario axis rewards.

\paragraph{Idea excerpts.} Question at most 180 characters, method at most 280; longer fields are cut. Judge comments are in the supplementary PDF.

\begin{sloppypar}\small
\noindent\textbf{CoI-Agent: ModalityDiff.} \textit{M.} ModalityDiff, a unified multi-scale SE(3)-equivariant continuous-discrete diffusion framework tailored for emerging drug modalities (RNA targets and peptidomimetic/PPI interfaces) equipped with Intrinsic Interface Preference Alignment: - Hierarchical Macromolecular Interface Enco [...]

\smallskip
\noindent\textbf{SciMON: TensorField-Diff.} \textit{Q.} How can 3D generative diffusion models simultaneously (1) satisfy directional quantum-mechanical/biophysical interaction constraints at catalytic or functionally critical residues, [...] \textit{M.} TensorField-Diff consists of three core architectural and algorithmic components: 1. Directional Biophysical Interaction Tensor Field ( mathbfT textres): 2. [...]

\smallskip
\noindent\textbf{AI-Researcher: DiffMOD.} \textit{Q.} Can pre-trained SE(3)-equivariant generative 3D diffusion trajectories be steered simultaneously toward biochemically critical target sub-pockets and optimal pharmacokinetic (PK/AD [...]

\smallskip
\noindent\textbf{Virtual-Scientists: Physically Calibrated Equivariant Diffusion for RNA-Targeted Ligand Design: [...].} \textit{M.} Targeting structured non-coding and regulatory RNA elements (e.g., riboswitches, viral internal ribosome entry sites, pre-miRNA hairpins, and retroviral frameshifting elements) with drug-like small molecules represents a major frontier in treating intractable genetic and infectio [...]

\smallskip
\noindent\textbf{ResearchAgent: We propose.} \textit{Q.} How can we develop a multi-objective, pocket-conditioned 3D diffusion framework for de novo macrocyclic peptide design that simultaneously optimizes structural 3D binding affinity/ [...] \textit{M.} We propose MacroDiff-ADMET, a multi-objective, mathrmSE(3)-equivariant geometric diffusion framework tailored for structure-based de novo design of macrocyclic peptides with co-optimized binding geometry and pharmacokinetic (ADMET) properties. [...]

\smallskip
\noindent\textbf{AI-Scientist-V1: PharmacoPocket-Diff.} \textit{Q.} How can 3D pocket-conditioned equivariant diffusion models be guided during reverse sampling to concurrently optimize binding affinity, geometric stereochemical validity, and multi [...]

\smallskip
\noindent\textbf{AgentLaboratory: PhysPK-Diff.} \textit{Q.} Can the simultaneous integration of differentiable internal-strain potentials and multi-task pharmacokinetic (PK/ADMET) value-function guidance during reverse SE(3)-equivariant dif [...] \textit{M.} PhysPK-Diff implements a dual-guided SE(3)-equivariant reverse diffusion framework built upon a pre-trained, parameter-frozen 3D pocket diffusion backbone: 1. [...]

\smallskip
\noindent\textbf{AI-Scientist-V2: GlueDiff.} \textit{Q.} Structure-based generative diffusion models (e.g., TargetDiff, DiffSBDD) fail when applied to molecular glue discovery because they treat the binding site as a single, deep, monoli [...] \textit{M.} GlueDiff, an SE(3)-equivariant continuous-discrete diffusion framework tailored for bipartite composite interfaces. GlueDiff represents the target-ligase interface as a multi-chain bipartite graph and introduces Bipartite Cross-Attention layers that explicitly propagate geometric [...]
\end{sloppypar}

\subsection{Survey: scientific discovery with AI}

\emph{survey\_50\_scientific\_hypothesis\_generation.}

The input is the survey \emph{Scientific Discovery in the Age of Artificial Intelligence} (Wang et al.). The packaged challenges concern scaling, closed-loop labs, and moving from literature mining to hypotheses that can be checked outside the training corpus.

\begin{table}[h]
\caption{All 12 dimension scores on \emph{survey\_50\_scientific\_hypothesis\_generation}. Highest in each row is bold.}
\label{tab:cs_survey50scientifichypothesisgeneration}
\begin{center}
\small
\setlength{\tabcolsep}{3.6pt}
\resizebox{\linewidth}{!}{%
\begin{tabular}{lcccccccc}
\toprule
Dimension & \shortstack{CoI-\\Agent} & SciMON & \shortstack{AI-\\Researcher} & \shortstack{Virtual-\\Scientists} & \shortstack{Research\\Agent} & \shortstack{AI-Scientist-\\V1} & \shortstack{Agent\\Laboratory} & \shortstack{AI-Scientist-\\V2} \\
\midrule
Context grounding & \textbf{5} & \textbf{5} & \textbf{5} & \textbf{5} & \textbf{5} & \textbf{5} & \textbf{5} & 4 \\
Problem formulation & \textbf{5} & \textbf{5} & \textbf{5} & \textbf{5} & \textbf{5} & \textbf{5} & \textbf{5} & \textbf{5} \\
Motivation consistency & \textbf{5} & \textbf{5} & \textbf{5} & \textbf{5} & \textbf{5} & \textbf{5} & \textbf{5} & \textbf{5} \\
Testability & \textbf{5} & \textbf{5} & \textbf{5} & 4 & 3 & \textbf{5} & \textbf{5} & 4 \\
Problem novelty & 3 & \textbf{4} & \textbf{4} & 3 & 3 & \textbf{4} & 3 & \textbf{4} \\
Method novelty & 3 & \textbf{4} & 3 & 3 & 3 & 3 & 2 & 3 \\
Scenario novelty & 3 & \textbf{4} & 3 & 3 & 2 & \textbf{4} & 1 & \textbf{4} \\
Insight novelty & 3 & \textbf{4} & 3 & 3 & 2 & \textbf{4} & 2 & 3 \\
Technical feasibility & \textbf{5} & \textbf{5} & 4 & \textbf{5} & \textbf{5} & 4 & 4 & 3 \\
Data feasibility & \textbf{5} & \textbf{5} & \textbf{5} & \textbf{5} & \textbf{5} & 4 & \textbf{5} & 4 \\
Budget feasibility & \textbf{5} & \textbf{5} & 4 & \textbf{5} & \textbf{5} & \textbf{5} & \textbf{5} & \textbf{5} \\
Verification feasibility & \textbf{5} & \textbf{5} & \textbf{5} & \textbf{5} & 3 & \textbf{5} & \textbf{5} & \textbf{5} \\
\bottomrule\end{tabular}%
}
\end{center}\end{table}
\paragraph{Why the scores move.}
{SciMON} is the only family at 4 on all four novelty axes and still at 5 on testability and technical feasibility. CONTRADICT-ICP grounds literature claims in LINCS and DepMap perturbational matrices and treats hypothesis search as invariant causal prediction; the judge reads that as a new measurement setting, not another LLM debate. {AI-Scientist-V1} is close (novelty 3.75, scenario 4) with a falsification-driven discovery loop. {AI-Scientist-V2} matches them on scenario (4) but drops to technical 3: CRUCIBLE leans on 70B-class APIs and underspecified simulators. {AgentLaboratory} is the floor (novelty 2.00, scenario 1): C2HS is conformal risk control on TDC and MatBench, a standard drug or materials surrogate loop. {ResearchAgent} names EpiLoop-SciAgent and stays on BindingDB and DUD-E; testability falls to 3 because the judge wants a tighter link from the OOD claim to a decision rule. Several families again converge on Popperian or conformal wrappers; only the systems that change what counts as evidence (perturbational screens, historical inflection experiments) move scenario.

\paragraph{Idea excerpts.} Question at most 180 characters, method at most 280; longer fields are cut. Judge comments are in the supplementary PDF.

\begin{sloppypar}\small
\noindent\textbf{CoI-Agent: Causal-SciAgent.} \textit{M.} Causal-SciAgent, an autonomous closed-loop scientific discovery framework grounded in Active Counterfactual Invariance Testing and Causal Graph-of-Trace (C-GoT) Auditing. 1. [...]

\smallskip
\noindent\textbf{SciMON: CONTRADICT-ICP.} \textit{Q.} How can we formally identify transferable, out-of-distribution scientific hypotheses from literature corpora without falling into correlation bias or unidentifiable pseudo-causalit [...] \textit{M.} CONTRADICT-ICP comprises four integrated phases: 1. Signed Context-Conditioned Literature Hypergraph (H context): Construct H = (V, E+, E-, C) from PubMed/PMC open-access literature up to a temporal cutoff (e.g., pre-2018), where entities V are grounded to NCBI Gene IDs / PubChem [...]

\smallskip
\noindent\textbf{AI-Researcher: PopperNet.} \textit{Q.} Can an adversarial hypothesis-generation framework - integrating typed mechanistic hypergraphs, physics-informed surrogate falsification, and split conformal prediction - guarantee fin [...]

\smallskip
\noindent\textbf{Virtual-Scientists: Calibrated Prospective Discovery.} \textit{M.} Calibrated Prospective Discovery (CPD), an open-source, mathematically grounded framework for debiased, causally robust, and prospectively calibrated scientific discovery. CPD operates via three core technical pillars: 1. [...]

\smallskip
\noindent\textbf{ResearchAgent: We propose.} \textit{Q.} How can we develop an uncertainty-aware, closed-loop hypothesis generation and verification framework for multi-agent automated scientific discovery that mitigates distribution shi [...] \textit{M.} We propose EpiLoop-SciAgent (Epistemic-Calibrated Closed-Loop Scientific Discovery Multi-Agent Framework), an uncertainty-aware, multi-agent architecture designed to mitigate out-of-distribution (OOD) hallucinations and bridge the gap between literature mining and experimental ve [...]

\smallskip
\noindent\textbf{AI-Scientist-V1: CounterSci.} \textit{Q.} Can formulating scientific hypotheses as interventional queries over epistemic structural causal hypergraphs, combined with active surrogate refutation, improve the discovery of tr [...] \textit{M.} The CounterSci framework operates in three sequential, closed-loop stages: (1) Epistemic SCM Extraction: Parse chronological corpora of peer-reviewed papers into an Epistemic Structural Causal Hypergraph mathcalH = ( mathcalV, mathcalE, Sigma), where nodes mathcalV [...]

\smallskip
\noindent\textbf{AgentLaboratory: C2HS.} \textit{Q.} How can we mathematically bound the failure rate and eliminate pathological proxy exploitation in LLM-driven scientific hypothesis generation under distribution shifts, while maint [...] \textit{M.} Conformalized Causal Hypothesis Search (C2HS) operates in three tightly coupled phases: 1) Literature Causal DAG Extraction: An open-source LLM extracts structured causal assertions from literature corpora to build a domain-specific causal DAG representing mechanistic dependencie [...]

\smallskip
\noindent\textbf{AI-Scientist-V2: CRUCIBLE.} \textit{Q.} Current LLM-based scientific discovery systems suffer from a severe alignment and evaluation gap: they optimize for narrative plausibility and literature recall, generating unfalsi [...] \textit{M.} CRUCIBLE, a framework and evaluation benchmark for falsifiability-driven scientific hypothesis discovery. CRUCIBLE reformulates hypothesis generation as an adversarial dialectic: given an empirical anomaly or domain context, the system generates a pair of mutually exclusive mecha [...]
\end{sloppypar}

\subsection{Workshop: symmetry and geometry in neural representations}

\emph{ws\_39\_neural\_representations\_geometry.}

The input is the NeurIPS 2024 CFP \emph{Workshop on Symmetry and Geometry in Neural Representations}. It asks how biological circuits and geometric deep learning preserve the geometry of the systems they represent, including approximate or broken symmetries.

\begin{table}[h]
\caption{All 12 dimension scores on \emph{ws\_39\_neural\_representations\_geometry}. Highest in each row is bold.}
\label{tab:cs_ws39neuralrepresentationsgeometry}
\begin{center}
\small
\setlength{\tabcolsep}{3.6pt}
\resizebox{\linewidth}{!}{%
\begin{tabular}{lcccccccc}
\toprule
Dimension & \shortstack{CoI-\\Agent} & SciMON & \shortstack{AI-\\Researcher} & \shortstack{Virtual-\\Scientists} & \shortstack{Research\\Agent} & \shortstack{AI-Scientist-\\V1} & \shortstack{Agent\\Laboratory} & \shortstack{AI-Scientist-\\V2} \\
\midrule
Context grounding & \textbf{5} & \textbf{5} & \textbf{5} & 4 & \textbf{5} & 1 & 4 & \textbf{5} \\
Problem formulation & \textbf{5} & \textbf{5} & \textbf{5} & \textbf{5} & \textbf{5} & 1 & \textbf{5} & \textbf{5} \\
Motivation consistency & \textbf{5} & \textbf{5} & \textbf{5} & \textbf{5} & \textbf{5} & 1 & 3 & \textbf{5} \\
Testability & \textbf{5} & \textbf{5} & \textbf{5} & 4 & 4 & 4 & 2 & 4 \\
Problem novelty & \textbf{4} & \textbf{4} & \textbf{4} & \textbf{4} & 3 & 2 & 3 & \textbf{4} \\
Method novelty & \textbf{4} & \textbf{4} & \textbf{4} & 3 & 3 & 2 & 1 & \textbf{4} \\
Scenario novelty & \textbf{4} & \textbf{4} & 1 & 3 & 1 & 1 & 2 & 2 \\
Insight novelty & \textbf{4} & \textbf{4} & \textbf{4} & \textbf{4} & 3 & 2 & 3 & \textbf{4} \\
Technical feasibility & \textbf{5} & \textbf{5} & \textbf{5} & 4 & 4 & 4 & 2 & 2 \\
Data feasibility & \textbf{5} & \textbf{5} & \textbf{5} & \textbf{5} & \textbf{5} & \textbf{5} & 2 & \textbf{5} \\
Budget feasibility & \textbf{5} & \textbf{5} & \textbf{5} & 4 & \textbf{5} & 2 & 2 & \textbf{5} \\
Verification feasibility & \textbf{5} & \textbf{5} & \textbf{5} & \textbf{5} & \textbf{5} & 4 & 1 & 4 \\
\bottomrule\end{tabular}%
}
\end{center}\end{table}
\paragraph{Why the scores move.}
This panel has the largest within-topic spread we plot. {CoI-Agent} and {SciMON} both sit at novelty 4.00 including scenario 4: they treat continuous attractors and principal bundles as objects that can be aligned between fly ellipsoid-body recordings, grid cells, and Neural ODEs, which is what the CFP asked for. {AI-Scientist-V2} has method and insight 4 but technical feasibility 2, because the Method field is a duplicate of Problem and Motivation and the atlas-bundle regularizer is never written down. {AI-Scientist-V1} collapses to novelty 1.75: it leaves the workshop and writes a chain-of-thought steering paper on AIME and MATH-500, so the judge marks scenario 1 and treats the problem as already studied in mechanistic interpretability. {AgentLaboratory} is worse on execution (testability 2, technical 2, budget 2): the GeoBreak draft cuts off at the method header, so the judge has almost nothing to score. {AI-Researcher} keeps method 4 on DLTA-Flow but scenario 1, because the evaluation stays on standard broken-symmetry forecasting rather than neural recordings. Geometry topics punish incomplete JSON harder than agentic-science topics, where a debate template still fills every field.

\paragraph{Idea excerpts.} Question at most 180 characters, method at most 280; longer fields are cut. Judge comments are in the supplementary PDF.

\begin{sloppypar}\small
\noindent\textbf{CoI-Agent: Gauge-Equivariant Dynamical Alignment and Flow Regularization for Continuous Neural [...].} \textit{M.} Let the internal state of a continuous-time recurrent network or Neural ODE be governed by: dotz(t) = F theta(z(t), u(t)), quad z in mathcalM subset mathbbR d, ; u in mathcalU where u(t) is an external control input (e.g., angular velocity). [...]

\smallskip
\noindent\textbf{SciMON: GaugeBundle-ODE.} \textit{Q.} How can an unsupervised deep generative dynamical model discover non-trivial latent manifold topologies (principal bundles with non-zero topological invariants) and continuous Lie [...] \textit{M.} GaugeBundle-ODE parameterizes the latent state space as a principal G-bundle P - B with the following mathematical components: 1. [...]

\smallskip
\noindent\textbf{AI-Researcher: alongside an.} \textit{Q.} How to address Workshop on Symmetry and Geometry in Neural Representations effectively under academic budget? \textit{M.} alongside an intrinsic Riemannian pull-back tangent bundle metric enables: - Robust generalization under symmetry-breaking perturbations. - Exact non-Euclidean metric alignment between artificial deep representations and biological neural population manifolds. [...]

\smallskip
\noindent\textbf{Virtual-Scientists: When Symmetries Break.} \textit{M.} Geometric Deep Learning (GDL) fundamentally relies on exact algebraic symmetry priors, constraining neural representations to be strictly invariant or equivariant under continuous transformation groups such as SO(3), SE(3), or O(3). [...]

\smallskip
\noindent\textbf{ResearchAgent: RNNs.} \textit{Q.} How can artificial recurrent neural networks (RNNs) and continuous attractor networks (CANs) be trained using geometric regularization (e.g., Lie-algebraic equivariance loss and pe [...] \textit{M.} Lie-Topological Attractor Regularization (LTAR) is a coordinate-free training framework for recurrent neural networks that induces continuous attractor manifolds matching target symmetric topologies without hardcoded architectural constraints. [...]

\smallskip
\noindent\textbf{AI-Scientist-V1: Detecting and Mitigating Reasoning Drift.} \textit{Q.} How to advance Workshop on Symmetry and Geometry in Neural Representations under academic budget? \textit{M.} 1. Benchmarks \& Models: Evaluate on DeepSeek-R1-Distill (8B, 14B, 32B) and QwQ-32B across challenging reasoning benchmarks (AIME 2024, MATH-500, OlympiadBench, LogiQA-2, and Codeforces). 2. [...]

\smallskip
\noindent\textbf{AgentLaboratory: exact Lie group equivariance.} \textit{Q.} How to address Workshop on Symmetry and Geometry in Neural Representations effectively under academic budget? \textit{M.} exact Lie group equivariance to robustly model physical dynamical systems and neural population codes subject to real-world anisotropic, symmetry-breaking perturbations? [...]

\smallskip
\noindent\textbf{AI-Scientist-V2: Spontaneous Fiber Bundle Factorization and Topological Obstructions in Latent [...].} \textit{Q.} Standard neural networks trained on data manifolds admitting continuous group actions (such as SO(2), SE(2), and SO(3)) spontaneously decompose latent representations into a princi [...] \textit{M.} Symmetry and geometry are foundational to both biological neural representations and geometric deep learning. While unconstrained deep neural networks often discover structured low-dimensional representations, the exact topological and fiber bundle characteristics of these latent [...]
\end{sloppypar}

\subsection{Workshop: generative AI and biology}

\emph{ws\_19\_generative\_biology.}

The input is the NeurIPS 2023 CFP \emph{Generative AI and Biology (GenBio)}. It concerns generative models that predict, characterize, and design biomolecules, with antibody sequence-structure co-design as the default attractive task.

\begin{table}[h]
\caption{All 12 dimension scores on \emph{ws\_19\_generative\_biology}. Highest in each row is bold.}
\label{tab:cs_ws19generativebiology}
\begin{center}
\small
\setlength{\tabcolsep}{3.6pt}
\resizebox{\linewidth}{!}{%
\begin{tabular}{lcccccccc}
\toprule
Dimension & \shortstack{CoI-\\Agent} & SciMON & \shortstack{AI-\\Researcher} & \shortstack{Virtual-\\Scientists} & \shortstack{Research\\Agent} & \shortstack{AI-Scientist-\\V1} & \shortstack{Agent\\Laboratory} & \shortstack{AI-Scientist-\\V2} \\
\midrule
Context grounding & \textbf{5} & \textbf{5} & \textbf{5} & \textbf{5} & \textbf{5} & \textbf{5} & \textbf{5} & \textbf{5} \\
Problem formulation & \textbf{5} & \textbf{5} & \textbf{5} & \textbf{5} & \textbf{5} & \textbf{5} & \textbf{5} & \textbf{5} \\
Motivation consistency & \textbf{5} & \textbf{5} & \textbf{5} & \textbf{5} & \textbf{5} & \textbf{5} & \textbf{5} & \textbf{5} \\
Testability & \textbf{5} & \textbf{5} & \textbf{5} & 4 & 4 & \textbf{5} & \textbf{5} & \textbf{5} \\
Problem novelty & 3 & 3 & 3 & \textbf{4} & 3 & 3 & 2 & \textbf{4} \\
Method novelty & 3 & \textbf{4} & 3 & 3 & 3 & 3 & 3 & \textbf{4} \\
Scenario novelty & 3 & 1 & 1 & \textbf{4} & 1 & 2 & 1 & 1 \\
Insight novelty & 3 & \textbf{4} & 3 & 3 & 2 & 3 & 3 & \textbf{4} \\
Technical feasibility & \textbf{5} & 4 & \textbf{5} & \textbf{5} & \textbf{5} & \textbf{5} & \textbf{5} & 4 \\
Data feasibility & \textbf{5} & \textbf{5} & \textbf{5} & \textbf{5} & \textbf{5} & \textbf{5} & \textbf{5} & \textbf{5} \\
Budget feasibility & 4 & \textbf{5} & \textbf{5} & 2 & 4 & 4 & \textbf{5} & 4 \\
Verification feasibility & \textbf{5} & \textbf{5} & \textbf{5} & \textbf{5} & \textbf{5} & \textbf{5} & \textbf{5} & \textbf{5} \\
\bottomrule\end{tabular}%
}
\end{center}\end{table}
\paragraph{Why the scores move.}
Virtual Scientists are the only family with scenario 4 and the novelty lead (3.50). Conformal-MPNN conditions inverse folding on conformational ensembles (MD, multi-state cryo-EM, NMR) rather than a static crystal, which the judge treats as a change of biological operating constraint. They pay for it in budget feasibility (2): ensemble featurization plus conformal calibration does not fit the 200 GPU hour envelope as written. {AI-Scientist-V2} and {SciMON} keep method 4 and stay at scenario 1, the same method versus setting split as on the tool survey: ChronoFlow and HoloFlow-Ab rewrite antibody CDR flow matching on SAbDab, but they do not leave antibody co-design. {ResearchAgent} (SE-GuidedOpt) and {AgentLaboratory} (ConfFlow-Ab) are the conservative inner ring (novelty 2.25): inference-time energy guidance or conformation-aware flow on the same CDR-H3 benches. {CoI-Agent} reaches scenario 3 with uncertainty-guided multi-conformational flow, short of the ensemble move. On biology CFPs, multi-persona discussion is what actually changes the setting; novelty boosting improves the generator inside the default antibody task.

\paragraph{Idea excerpts.} Question at most 180 characters, method at most 280; longer fields are cut. Judge comments are in the supplementary PDF.

\begin{sloppypar}\small
\noindent\textbf{CoI-Agent: ConformFlow-UG.} \textit{M.} ConformFlow-UG addresses these bottlenecks by introducing an ensemble-conditioned equivariant flow matching paradigm with epistemic uncertainty-penalized vector-field guidance. [...]

\smallskip
\noindent\textbf{SciMON: HoloFlow-Ab.} \textit{Q.} Can we formulate antibody CDR-H3 generative co-design directly on a kinematically closed Riemannian sub-manifold coupled with an asymmetric induced-fit vector field on the antigen [...]

\smallskip
\noindent\textbf{AI-Researcher: EquiFlow-Steer.} \textit{Q.} How can we achieve unified, sample-efficient joint generation of 3D backbone conformations (SE(3) L) and discrete amino acid sequences (S in A L) for flexible antibody CDR loops su [...] \textit{M.} EquiFlow-Steer integrates three core components: (1) Unified Multi-Modal Riemannian Flow Matching on SE(3) L x Delta 19L parameterizing translational C alpha paths, rotational SO(3) Lie-algebra geodesic paths, and continuous sequence logit simplex flows. [...]

\smallskip
\noindent\textbf{Virtual-Scientists: Conformal-MPNN.} \textit{M.} Conformal-MPNN, an uncertainty-calibrated generative inverse folding framework that operates directly over dynamic conformational ensembles. Conformal-MPNN introduces two synergistic technical innovations: 1. [...]

\smallskip
\noindent\textbf{ResearchAgent: Structure Projection.} \textit{Q.} How can we formulate an inference-time, plug-and-play guidance framework that couples frozen structural oracle feedback, such as ESMFold or AlphaFold confidence and biophysical con [...] \textit{M.} `SE-GuidedOpt` is an inference-time, plug-and-play guidance framework that operates directly on pre-trained sequence-structure antibody co-diffusion models (e.g., DiffAb or RFdiffusion-AA conditioned on an epitope) without requiring model retraining or fine-tuning. [...]

\smallskip
\noindent\textbf{AI-Scientist-V1: Ensemble-Guided Flow Matching for Dynamic Sequence-Structure Co-Design in Antibody [...].} \textit{Q.} Can SE(3)-equivariant joint discrete-continuous flow matching conditioned on conformational ensembles and guided by evolutionary priors generate antibody CDR loops with superior dy [...] \textit{M.} 1. Joint Sequence-Structure Flow Matching: Formulate antibody CDR co-design as a probability flow on the product manifold M = Delta (L x 20) x SE(3) L, where Delta represents the probability simplex over 20 amino acids and SE(3) represents residue rigid frames (Calpha position x i in R 3 and [...]

\smallskip
\noindent\textbf{AgentLaboratory: ConfFlow-Ab.} \textit{Q.} Can a lightweight, continuous SE(3) x Delta 19 Riemannian Flow Matching adapter conditioned on frozen foundational protein embeddings jointly inpaint antibody CDR-H3 backbone geome [...] \textit{M.} ConfFlow-Ab introduces a Riemannian Flow Matching (RFM) architecture over the product manifold M = (R 3 x SO(3) x Delta 19) L for antibody CDR-H3 inpainting: 1. [...]

\smallskip
\noindent\textbf{AI-Scientist-V2: ChronoFlow.} \textit{Q.} Joint sequence-structure flow matching models enforce a rigid synchronous time schedule t in [0, 1] across continuous backbone SE(3) geometries and discrete categorical sequence sp [...] \textit{M.} ChronoFlow, an asynchronous multi-scale flow matching framework for antibody sequence-structure co-design. ChronoFlow decouples sequence and structural flow velocities via a biophysically grounded, adaptive temporal warping schedule tau seq(t) = tau struct(t) modulated by the in [...]
\end{sloppypar}

\subsection{Survey: causal representation learning}

\emph{survey\_46\_causal\_representation\_learning.}

The input is the survey \emph{Toward Causal Representation Learning} (Sch\"olkopf et al.). The packaged gap is identifiability: recovering high-level causal variables and mechanisms from low-level observations, often under unpaired or imperfect interventions.

\begin{table}[h]
\caption{All 12 dimension scores on \emph{survey\_46\_causal\_representation\_learning}. Highest in each row is bold.}
\label{tab:cs_survey46causalrepresentationlearning}
\begin{center}
\small
\setlength{\tabcolsep}{3.6pt}
\resizebox{\linewidth}{!}{%
\begin{tabular}{lcccccccc}
\toprule
Dimension & \shortstack{CoI-\\Agent} & SciMON & \shortstack{AI-\\Researcher} & \shortstack{Virtual-\\Scientists} & \shortstack{Research\\Agent} & \shortstack{AI-Scientist-\\V1} & \shortstack{Agent\\Laboratory} & \shortstack{AI-Scientist-\\V2} \\
\midrule
Context grounding & \textbf{5} & \textbf{5} & \textbf{5} & \textbf{5} & \textbf{5} & \textbf{5} & 4 & 4 \\
Problem formulation & \textbf{5} & \textbf{5} & \textbf{5} & \textbf{5} & \textbf{5} & \textbf{5} & \textbf{5} & \textbf{5} \\
Motivation consistency & \textbf{5} & \textbf{5} & \textbf{5} & \textbf{5} & \textbf{5} & \textbf{5} & \textbf{5} & \textbf{5} \\
Testability & \textbf{5} & \textbf{5} & \textbf{5} & 4 & 4 & \textbf{5} & \textbf{5} & 4 \\
Problem novelty & \textbf{4} & \textbf{4} & 3 & \textbf{4} & 3 & 3 & 3 & 3 \\
Method novelty & \textbf{4} & \textbf{4} & \textbf{4} & \textbf{4} & \textbf{4} & \textbf{4} & 3 & 3 \\
Scenario novelty & \textbf{4} & 3 & 2 & 3 & 1 & 1 & 1 & 2 \\
Insight novelty & \textbf{4} & \textbf{4} & \textbf{4} & \textbf{4} & 3 & \textbf{4} & 3 & \textbf{4} \\
Technical feasibility & \textbf{5} & \textbf{5} & \textbf{5} & 4 & \textbf{5} & \textbf{5} & 4 & 3 \\
Data feasibility & \textbf{5} & \textbf{5} & \textbf{5} & \textbf{5} & \textbf{5} & \textbf{5} & \textbf{5} & \textbf{5} \\
Budget feasibility & \textbf{5} & 4 & \textbf{5} & \textbf{5} & \textbf{5} & \textbf{5} & \textbf{5} & 4 \\
Verification feasibility & \textbf{5} & \textbf{5} & \textbf{5} & \textbf{5} & \textbf{5} & \textbf{5} & \textbf{5} & 4 \\
\bottomrule\end{tabular}%
}
\end{center}\end{table}
\paragraph{Why the scores move.}
{CoI-Agent} is again the only scenario 4 idea and the novelty leader (4.00). ICT-CRL puts identifiability in continuous time with irregular sampling (Causal-Pendulum, Causal3DIdent video, MuJoCo), which the judge reads as a new observation model rather than another autoencoder plus NOTEARS. {SciMON} and Virtual Scientists sit just behind (novelty 3.75, scenario 3) with score decomposition and imperfect-intervention robustness. {ResearchAgent} (SICA-VMR) and {AI-Scientist-V1} (latent invariant score matching) keep method 4 and scenario 1: they write a stronger identifiability gadget on Causal3DIdent, MPI3D, and Perturb-seq, without leaving the survey's standard multi-environment image setup. {AgentLaboratory} stays at novelty 2.50 with a contractive autoencoder plus DAGMA. {AI-Scientist-V2} has insight 4 but technical 3, another underspecified identifiability proof sketch. The pattern matches the tool-learning survey: debate and retrieval ({CoI-Agent}) change the constraint; boosting and lab-style agents improve the estimator inside the given CRL benches.

\paragraph{Idea excerpts.} Question at most 180 characters, method at most 280; longer fields are cut. Judge comments are in the supplementary PDF.

\begin{sloppypar}\small
\noindent\textbf{CoI-Agent: ICT-CRL.} \textit{M.} Let X(t) in mathbbR D be a high-dimensional observation trajectory (e.g., video frames) sampled at arbitrary, irregularly spaced timestamps t in [0, T], generated from underlying latent causal states Z(t) in mathbbR d (d ll D) via a smooth, non-linear mixing fun [...]

\smallskip
\noindent\textbf{SciMON: CasCADe-CRL.} \textit{Q.} How can non-linear high-dimensional causal representations and their underlying structural causal mechanisms be provably identified under continuous, unknown, and cascading soft di [...]

\smallskip
\noindent\textbf{AI-Researcher: L-SIEM.} \textit{Q.} Can we achieve provable, component-wise identifiability of non-additive, continuous latent Structural Causal Models (SCMs) from high-dimensional observational mixtures across heter [...] \textit{M.} L-SIEM combines a deterministic autoencoder backbone with an implicit energy-based latent causal model optimized via sliced score matching and sparse invariant mechanism regularization. The architecture comprises: 2. [...]

\smallskip
\noindent\textbf{Virtual-Scientists: RICRL.} \textit{M.} RICRL (Robust Identifiable Causal Representation Learning) resolves this critical challenge via three fundamental advancements: 1. [...]

\smallskip
\noindent\textbf{ResearchAgent: The refined.} \textit{Q.} How can we guarantee the identifiable disentanglement and recovery of latent causal mechanisms from high-dimensional observational and interventional data when interventional targe [...] \textit{M.} Identifiable Sparse-Invariant Causal Autoencoding with Variational Mechanism-Target Routing (SICA-VMR) The refined SICA-VMR framework establishes a complete, closed-form, and theoretically guaranteed solution for learning identifiable nonlinear causal representations and structur [...]

\smallskip
\noindent\textbf{AI-Scientist-V1: Latent Invariant Score Matching.} \textit{Q.} Can non-linear causal representations and their underlying Structural Causal Model (SCM) be provably identified up to permutation and element-wise diffeomorphism from high-dimensio [...] \textit{M.} Latent Invariant Score Matching (LISM) operates via three interconnected components: 1. Identifiability Formalism: We prove that if interventions act sparsely on a subset of causal mechanisms across regimes, the latent score difference vectors Delta s(z) = nabla z log p e 1 [...]

\smallskip
\noindent\textbf{AgentLaboratory: Lossy-CRL.} \textit{Q.} Can we provably and empirically achieve identifiable causal representation learning from high-dimensional visual observations without requiring exact bijectivity (flows) or known s [...] \textit{M.} Lossy-CRL consists of: (1) A contractive deterministic autoencoder (Encoder f theta: R D - R d, Decoder g phi: R d - R D) regularized by the Frobenius norm of the encoder Jacobian to guarantee isometric embedding; (2) A multi-environment score estimator s psi(z, e) trained via [...]

\smallskip
\noindent\textbf{AI-Scientist-V2: Minimal Interventional Footprint.} \textit{Q.} Minimal Interventional Footprint (MIF) - which minimizes the coordinate sparsity of latent transition residuals alongside generative reconstruction - is sufficient to provably identify [...] \textit{M.} Minimal Interventional Footprint (MIF) framework for causal representation learning in continuous dynamical systems. We prove that when actions perturb an unknown, sparse subset of latent causal mechanisms (the Sparse Mechanism Shift principle), regularizing the ell 1-norm/ent [...]
\end{sloppypar}